\documentclass[a4paper,fleqn]{cas-sc}
\usepackage[labelfont=bf,labelsep=period]{caption}
\usepackage[authoryear]{natbib}
\usepackage{subcaption} %子图支持
\newcommand{\rev}[1]{#1}

\def\tsc#1{\csdef{#1}{\textsc{\lowercase{#1}}\xspace}}
\tsc{WGM}
\tsc{QE}
\tsc{EP}
\tsc{PMS}
\tsc{BEC}
\tsc{DE}
\begin{document}

	\let\WriteBookmarks\relax
	\def\floatpagepagefraction{1}
	\def\textpagefraction{.001}
	\shorttitle{Weakly Supervised WSI Classification with DSAGL}
	\shortauthors{Cao et al.}
%	\begin{frontmatter}
	
	\title [mode = title]{Dual-stream attention-guided learning for weakly supervised whole slide image classification}
	
	% ========================= 
	% Authors 
	% =========================
	\author[1]{Daoxi Cao}[orcid=0009-0003-7157-4802]
	\fnmark[1]
	\ead{caodaoxi6432@link.tyut.edu.cn}
	
	\author[1]{Hangbei Cheng}[orcid=0009-0006-7310-7636]
	\fnmark[1]
	\ead{chenghangbei0702@163.com}
	
	\author[2]{Yijin Li}[orcid=0009-0008-3418-0253]
	\ead{liyijin4335@link.tyut.edu.cn}
	
	\author[1]{Ruolin Zhou}[orcid=0009-0007-6038-7170]
	\ead{zhouruolin4817@link.tyut.edu.cn}
	
	\author[1]{Xuehan Zhang}[orcid=0009-0001-3734-8874]
	\ead{zhangxuehan7154@link.tyut.edu.cn}
	
	\author[3]{Xinyi Li}[orcid=0009-0006-3743-8411]
	\ead{lixinyi7064@link.tyut.edu.cn}
	
	\author[4]{Binwei Li}[orcid=0009-0007-7263-7720]
	\ead{libinwei7791@link.tyut.edu.cn}
	
	\author[4]{Xuancheng Gu}[orcid=0009-0009-8980-434X]
	\ead{guxuancheng@bupt.edu.cn}
	
	\author[5]{Jiannan Zhang}[orcid=0009-0001-0147-0990]
	\ead{zhangjiananm0@sina.com}
	
	\author[3]{Xueyu Liu}[orcid=0000-0001-5745-4722]
	\cormark[1]
	\ead{liuxueyu0229@link.tyut.edu.cn}
	
	\author[1]{Yongfei Wu}[orcid=0000-0002-5010-2561]
	\cormark[1]
	\ead{wuyongfei@tyut.edu.cn}
	
	% ========================= 
	% Affiliations 
	% =========================
	\affiliation[1]{ organization={College of Computer Science and Technology, College of Data Science, Taiyuan University of Technology}, city={Taiyuan}, postcode={030024}, state={Shanxi}, country={China} } 
	
	\affiliation[2]{ organization={College of Humanities, Law and Foreign Languages, Taiyuan University of Technology}, city={Taiyuan}, postcode={030024}, state={Shanxi}, country={China} } 
	
	\affiliation[3]{ organization={College of Artificial Intelligence, Taiyuan University of Technology}, city={Taiyuan}, postcode={030024}, state={Shanxi}, country={China} } 
	
	\affiliation[4]{ organization={School of Cyberspace Security, Beijing University of Posts and Telecommunications}, city={Beijing}, postcode={102206}, country={China} } 
	
	\affiliation[5]{ organization={School of Mathematics, Taiyuan University of Technology}, city={Taiyuan}, postcode={030024}, state={Shanxi}, country={China} } 
	
	% ========================= 
	% Footnotes 
	% =========================
	
	\cortext[1]{Xueyu Liu and Yongfei Wu are Corresponding authors.} 
	\fntext[1]{These authors contributed equally to this work.}
	
	\begin{abstract}	
	Whole slide images (WSIs) play a crucial role in cancer diagnosis due to their ultra-high resolution and rich morphological information, and multiple instance learning (MIL) has become a prevalent paradigm to solve the massive size of WSIs and the scarcity of fine-grained annotations of instance. However, most existing MIL methods struggle to accurately identify diagnostically critical local regions (instance) using only slide-level labels, and suffer from modelling the relationship of instances efficiently. To address these defects, we propose a Dual-Stream Attention-Guided Learning (DSAGL) framework. DSAGL bridges slide-level supervision and instance-level learning through a teacher-student dual-stream architecture, and mitigates instance ambiguity by generating attention-guided pseudo labels. The framework employs a shared lightweight encoder to efficiently model long-range dependencies and an attention-based fusion mechanism to enhance sensitivity to sparse, informative regions. Extensive experiments on synthetic benchmarks and real-world pathological WSI datasets demonstrate that DSAGL consistently outperforms state-of-the-art MIL methods, achieving superior discriminative performance and robustness under weak supervision.

	\end{abstract}
	
	\begin{comment}
		\begin{highlights}
			
			\item We propose DSAGL, a novel weakly supervised classification framework that integrates a dual-stream structure and a teacher–student mechanism to jointly enhance instance-level and bag-level performance.
			
			\item An alternating training strategy is introduced to improve semantic consistency and enable effective collaboration between the teacher and student branches.
			
			\item We design a lightweight encoder (VSSMamba) and a scale-aware attention module (FASA) to balance efficient long-range modeling and focus on diagnostically critical regions.
			
			\item DSAGL consistently outperforms representative MIL-based methods on both synthetic and real-world pathological datasets at the instance and bag levels.
			
		\end{highlights}
	\end{comment}
%\end{frontmatter}
	
	\maketitle
	
	\section{Introduction}
	%%1
	\rev{Whole slide images (WSIs) have become a fundamental imaging modality in modern pathological diagnosis due to their ultra-high resolution and rich morphological information~\citep{niazi2019digital,shmatko2022artificial,dimitriou2019deep,wu2024pan}. By capturing fine-grained tissue structures at the microscopic level, WSIs provide critical evidence for cancer diagnosis and treatment planning. However, their enormous size and complex tissue heterogeneity pose significant challenges for automated analysis~\citep{farahani2015whole,song2023artificial}. In particular, acquiring exhaustive instance-level annotations for WSIs requires substantial effort from expert pathologists, making fully supervised learning approaches impractical in large-scale clinical settings. These limitations have motivated increasing interest in weakly supervised learning paradigms for computational pathology~\citep{li2023task,lin2023interventional,nakhli2023sparse,ryu2023ocelot,tang2023multiple}.}
	
	%%2
	\rev{Among weakly supervised approaches, multiple instance learning (MIL) has emerged as a dominant framework for WSI classification~\citep{hou2022h,ilse2018attention,li2021dual,lu2021data}. In MIL, a WSI is treated as a bag containing numerous image patches, while only bag-level labels are available for training~\citep{zhang2022dtfd,amores2013multiple}. This formulation significantly reduces annotation costs and aligns well with clinical diagnostic practice. Despite these advantages, existing MIL methods still face notable limitations. Many approaches rely on simplistic aggregation strategies, such as attention-based pooling~\citep{ilse2018attention}, or implicit independence assumptions among instances, which restrict their ability to capture complex spatial dependencies across distant tissue regions. As a result, these models often struggle to accurately identify diagnostically relevant instances in scenarios characterized by sparse and heterogeneous pathological patterns. Moreover, under weak supervision, improving instance-level discriminability frequently comes at the expense of bag-level classification performance, limiting generalization and clinical applicability.}
	
	%%3
	\rev{Recent studies have explored more expressive architectures to address these limitations. Convolutional neural networks (CNNs)~\citep{alex2011imagenet} are constrained by their limited receptive fields, making it difficult to model long-range dependencies~\citep{gu2023mamba,hao2024t,vim}. Transformer-based architectures~\citep{han2022survey,liu2022video}, while capable of capturing global context, suffer from quadratic computational complexity with respect to the number of instances, resulting in poor scalability for gigapixel WSIs~\citep{vim,hao2024t}. In addition, widely adopted single-stream MIL architectures~\citep{jiang2022bimetallic} often lack explicit mechanisms for instance-level disambiguation, causing attention mechanisms to focus on visually dominant but diagnostically irrelevant regions. These challenges highlight the need for more efficient and robust weakly supervised frameworks that can jointly model long-range dependencies and improve instance-level discrimination.}
	
\begin{figure}[h]
		\centering
		\includegraphics[width=\linewidth]{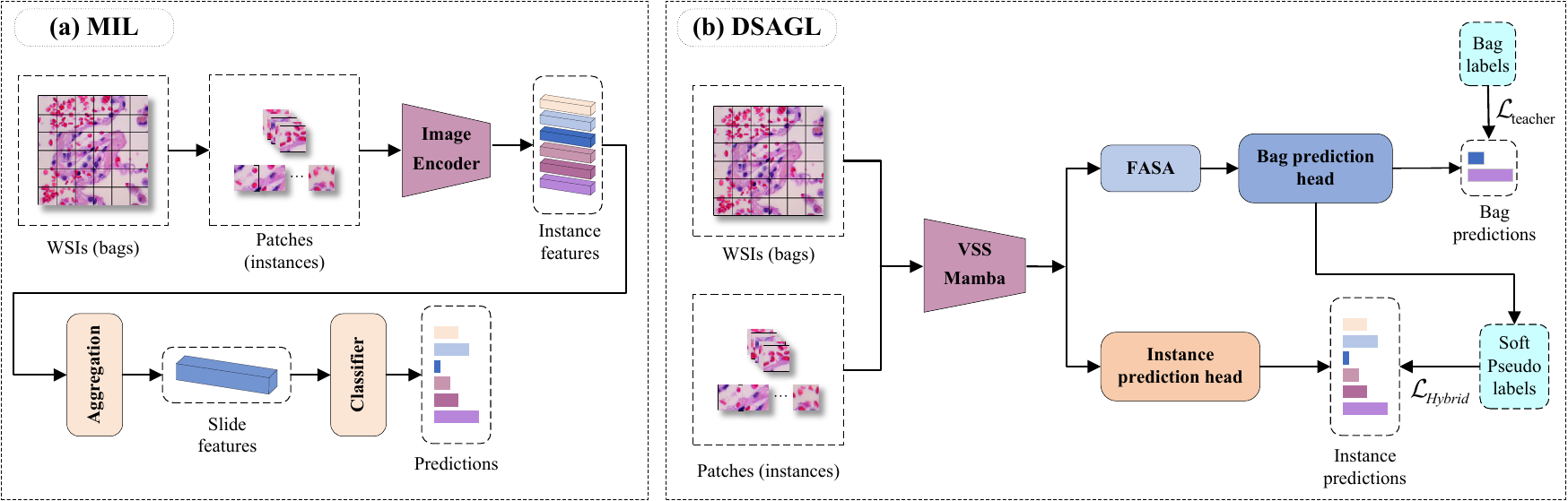}
		\caption{(a) Basic MIL architecture; (b) Our proposed DSAGL architecture.}
		\label{fig:compare_mil_dawn}
	\end{figure}

	%%4
	To address the aforementioned challenges, we propose DSAGL (Dual-Stream Attention-Guided Learning), a novel weakly supervised classification framework based on a dual-stream, teacher-student architecture with attention guidance. DSAGL is designed to capture inter-instance correlations while jointly enhancing both instance-level discriminability and bag-level classification performance in MIL models. The framework consists of a unified feature extraction backbone and a dual-stream structure. Specifically, the teacher stream employs a multi-scale attention mechanism to focus on diagnostically relevant instances, enabling accurate bag-level classification while generating high-quality soft pseudo labels. Guided by these pseudo labels, the student stream performs instance-level classification, thereby improving fine-grained discriminative capability. Moreover, as shown in Fig.~\ref{fig:compare_mil_dawn}(b), DSAGL employs an alternating training strategy between the teacher and student \rev{streams}, in contrast to the sequential paradigm adopted by conventional dual-stream models. This collaborative optimization enhances semantic consistency and further improves the overall model performance.
	
	To further enhance the model's representational capacity and generalization under weak supervision, DSAGL incorporates two key components: a Vision-aware Skip-connected Selective Mamba encoder (VSSMamba) for efficient long-range dependency modeling with linear complexity, and a Fusion-Attentive Scale-Aware module (FASA) to emphasize sparse but diagnostically relevant regions. These designs enable DSAGL to effectively leverage coarse-grained labels for accurate WSI classification. Extensive experiments conducted on both two synthetic datasets (CIFAR-10 and NCT-CRC) and one real-world pathological dataset (TCGA-Lung) demonstrate that DSAGL outperforms representative MIL-based approaches at both instance and bag levels, highlighting its strong discriminative power and generalization across diverse scenarios.
	
	The main contributions of this work are summarized as follows:
	
	\begin{itemize}
		\item We propose a novel weakly supervised classification framework, DSAGL, which integrates a dual-stream architecture with a teacher-student mechanism to jointly optimize instance-level discriminability and bag-level classification performance, addressing the limitations of conventional MIL methods in semantic modeling.
		
		\item We introduce an alternating training strategy to enhance semantic consistency, enabling collaborative optimization between the teacher and student streams, which significantly improves the model's robustness and generalization in complex lesion recognition scenarios.
		
		\item  A lightweight backbone VSSMamba for efficient long-range dependency modeling and a multi-scale attention module FASA for capturing discriminative patterns in sparse tumor regions.
		
	\end{itemize}
	
	The remainder of this paper is organized as follows: Section 2 reviews related work and highlights current limitations and challenges. Section 3 focuses on the proposed model architecture, detailing its design and key technical components. Section 4 presents experimental results on three datasets along with ablation studies. Section 5 discusses the strengths and limitations of the proposed approach. Finally, Section 6 concludes the paper and outlines potential directions for future research.

	\section{Related Work}
	\subsection{Multiple Instance Learning for WSI Classification}
	
	Multiple instance learning (MIL)~\citep{wan2019c,shi2024vila,shi2024vila} has become a widely adopted weakly supervised paradigm in computational pathology, particularly well-suited for WSIs, which are characterized by extremely high resolution and large-scale tissue content. In MIL, each WSI is divided into a set of patches (instances) grouped into a bag, with supervision provided only at the bag level. This eliminates the need for dense annotations and significantly reduces manual labeling costs.
	
	Early MIL methods, such as MI-Net~\citep{wang2018revisiting}, employed max or mean pooling to aggregate instance features for bag-level prediction. While computationally efficient, these approaches are limited in their ability to distinguish semantically relevant instances. To address this, Ilse et al.~\citep{ilse2018attention} introduced Attention-MIL, which incorporates learnable attention weights to assign importance scores to instances. The framework was further extended to ABMIL by incorporating gated attention mechanisms, thereby improving both aggregation flexibility and interpretability.
	~\citep{ilse2018attention}
	Subsequent works have explored more expressive instance-level modeling strategies. Chen et al.~\citep{chen2020msa} proposed MSA-MIL by integrating multi-scale feature extraction modules, enhancing sensitivity to fine-grained structural variations. Li et al.~\citep{li2023weakly} introduced SA-MIL, which embeds self-attention into the aggregation process to capture contextual dependencies among instances. DTFD-MIL~\citep{zhang2022dtfd} employs a dual-path differential feature design to strengthen the model's ability to detect heterogeneous lesion patterns. In addition, DSMIL~\citep{li2021dual} introduces an explicit instance selection module within a \rev{dual-stream} structure, enabling collaborative instance- and bag-level decision-making. CLAM~\citep{lu2021data} incorporates attention pruning to localize diagnostically relevant regions and improve interpretability. Transformer-based frameworks such as TransMIL~\citep{shao2021transmil} further extend MIL by modeling global dependencies across instances, exhibiting superior long-range reasoning compared to traditional attention-based models.
	
	Despite these advances, two key limitations remain in current MIL-based approaches. First, most models rely solely on bag-level supervision and lack explicit instance-level guidance, which hinders precise localization of critical pathological regions. Second, attention mechanisms are typically used only for feature aggregation and are not explicitly aligned with classification objectives, limiting their effectiveness under weak supervision.

	\subsection{Dual-Stream Architectures and Knowledge Distillation}
	
	Prior to the emergence of dual-stream architectures, extensive research explored the integration of knowledge distillation (KD) into single-stream or loosely coupled frameworks to enhance the representational capacity and training stability of student models. For example, Co-MIL~\citep{jiang2022bimetallic} employs a multi-teacher collaborative distillation strategy, where separate teacher models extract global and local semantic information to guide the student through cross-model supervision. GLMC~\citep{du2023global} introduces bidirectional guidance between global and local streams, thereby improving the model's ability to capture multi-scale lesion features. Self-Distillation utilizes the historical states of the student model as an implicit teacher, enabling knowledge transfer without additional parameter overhead-an approach particularly suitable for resource-constrained settings. ReNA-KD~\citep{zhang2022region} incorporates region-structured attention maps into the distillation process, enhancing spatial awareness and improving recognition of heterogeneous lesion patterns.
	
	Building upon these foundational works, recent studies have further integrated knowledge distillation into dual-stream frameworks to enhance cross-stream prediction consistency and collaboration. R$^2$T-MIL~\citep{tang2024feature} introduces a region-aware token distillation module within a Transformer-based architecture, leveraging spatial attention to guide the student in focusing on diagnostically relevant areas. GLMKD~\citep{cheng2025glmkd} designs a dual-teacher framework with global and local streams, and incorporates a shape transfer loss to reconcile inconsistencies between pseudo labels from different teachers, enabling more accurate boundary modeling and improved training stability under weak supervision.
	
	While recent methods have introduced improvements, many distillation strategies remain constrained by static soft labels and rigid optimization phases. The absence of adaptive feedback mechanisms hampers their capacity to cope with semantic drift and evolving \rev{inter-stream} representations during training.
	
	\subsection{Long-Range Dependency Modeling in WSI}
	
	WSIs are composed of thousands of image patches and exhibit strong non-local structural characteristics. In pathological diagnosis, spatially distant regions often present semantic dependencies-for example, between tumor boundaries and cores, or between inflamed tissues and surrounding microenvironments. Modeling such long-range dependencies is critical for improving classification accuracy.
	
	Convolutional neural networks (CNNs)~\citep{alzubaidi2021review,chua1993cnn} are inherently limited by their local receptive fields and struggle to capture global contextual information. To address this, Transformer-based~\citep{vaswani2017attention,han2021transformer}  architectures have been introduced. TransMIL~\citep{shao2021transmil} is one of the earliest attempts to model inter-patch relationships via self-attention for global bag representation. Vision Transformer (ViT)~\citep{dosovitskiy2020image} extends standard Transformer designs to image modeling and has been widely adopted in MIL settings. DTFD-MIL further incorporates feature differencing to enhance the modeling of local-global disparities.
	
	Despite their expressive capacity, Transformer models suffer from high computational cost. The attention mechanism scales quadratically with input size, making it inefficient for WSIs with thousands of patches. This leads to memory bottlenecks and latency issues, hindering real-world deployment. To address these limitations, Structured State Space Models (SSMs)~\citep{gu2021efficiently} have been proposed as efficient alternatives. S4, introduced by Gu et al., first incorporated continuous-time state-space equations into deep networks for efficient long-range sequence modeling. Building on this, Mamba introduced selective state updates and input gating mechanisms, enabling fine-grained control over information flow while preserving the efficiency of SSMs. Mamba~\citep{gu2023mamba,hao2024t,ma2024u} achieves superior parallelism and scalability, outperforming standard Transformers in tasks such as image classification and semantic segmentation.
	
	Derived from the Mamba framework, several variants have been developed. Vision Mamba~\citep{vim} integrates convolutional pre-processing and spatial shift mechanisms to enhance spatial awareness. T-Mamba~\citep{hao2024t} introduces temporal decoupling for efficient modeling of non-uniform sequential data. U-Mamba~\citep{ma2024u} extends the architecture for biomedical image segmentation by incorporating a U-shaped encoder-decoder design with Mamba blocks, enabling long-range context modeling while preserving spatial resolution
	\section{Method}
	This section provides a detailed exposition of the proposed methodological framework, encompassing the formal definition of the problem, the overall architectural design of the model, and the construction principles of key components, including the VSSMamba and the Dual-Stream Network.
	\subsection{Task Definition}
	\subsubsection{Multiple Instance Learning (MIL)}
\begin{figure}[h]
		\centering
		\includegraphics[width=0.7\linewidth]{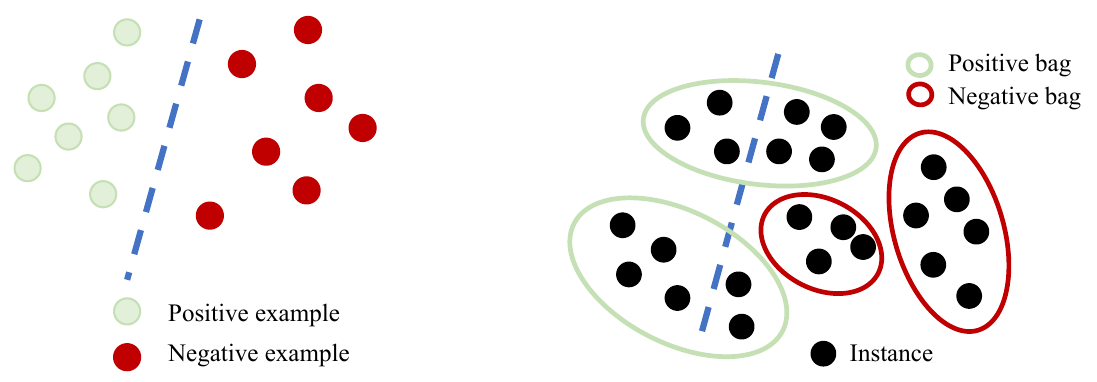}
		\caption{Supervised learning and Multiple-Instance Learning (MIL).}
		\label{fig:MIL}
	\end{figure}

	We consider weakly supervised pathological image classification under the Multiple Instance Learning (MIL) framework. As illustrated in Fig.~\ref{fig:MIL}, unlike traditional supervised learning where each instance is individually labeled, MIL assigns labels only at the bag level, with each bag containing multiple unlabeled instances.
	
	Each WSI is divided into a bag of image patches, where each patch is treated as an instance~\citep{ilse2018attention,li2023task,lin2023interventional,shi2023structure}. A bag is defined as \( B_i = \{x_{i,1}, \dots, x_{i,N_i}\} \), where \( x_{i,j} \in \mathbb{R}^{\texttt{d}} \) denotes the \( \texttt{d} \)-dimensional feature vector of the \( j \)-th instance in bag \( i \), and \( N_i \) is the number of instances.
	
	The dataset is represented as \( \mathcal{D} = \{(B_i, y_i)\}_{i=1}^M \), where only bag-level labels \( y_i \in \{0,1\} \) are available during training. According to the standard MIL assumption, a bag is positive if it contains at least one positive instance:
	\begin{equation}
		y_i = \max_{1 \leq j \leq N_i} z_{i,j},
	\end{equation}
	where \( z_{i,j} \in \{0,1\} \) denotes the (latent) instance label. The goal is to jointly optimize bag-level classification and instance-level discrimination under weak supervision.
	\subsubsection{SSM and Mamba}
	Structured State Space Models (SSMs)~\citep{vim} offer an efficient module for long-range sequence modeling with linear time complexity. Given a 1D input sequence \(x(t)\), the system evolves through a hidden state \(h(t) \in \mathbb{R}^\texttt{N}\), where \(\texttt{N}\) denotes the dimension of the hidden state, governed by the continuous-time dynamics:	
	\begin{equation}
		\frac{d}{dt} h(t) = \mathbf{A} h(t) + \mathbf{B} x(t), \quad y(t) = \mathbf{C} h(t),
	\end{equation}
	where \(\mathbf{A}\in \mathbb{R}^{\texttt{N} \times \texttt{N}}, \mathbf{B}\in \mathbb{R}^{\texttt{N} \times \texttt{1}}, \mathbf{C}\in \mathbb{R}^{\texttt{1} \times \texttt{N}}\) are learnable matrices. For discrete-time processing, we apply zero-order hold (ZOH) with time step \(\Delta\), yielding:
	\begin{equation}
		\mathbf{\overline{A}} = \exp(\Delta \mathbf{A}), \label{eq:discrete_a}
	\end{equation}
	\vspace{-0.3cm}
	\begin{equation}
		\mathbf{\overline{B}} = \left(\Delta \mathbf{A}\right)^{-1} \left( \exp(\Delta \mathbf{A}) - \mathbf{I} \right) \Delta \mathbf{B}, \label{eq:discrete_b}
	\end{equation}
	\vspace{-0.3cm}
	\begin{equation}
		h_t = \mathbf{\overline{A}} h_{t-1} + \mathbf{\overline{B}} x_t, \quad y_t = \mathbf{C} h_t. \label{eq:discrete_ht}
	\end{equation}
	
	The output sequence can also be expressed as a global convolution over a structured kernel:
	\begin{equation}
		\mathbf{\overline{K}} = \left( \mathbf{C} \mathbf{\overline{A}}^i \mathbf{\overline{B}} \right)_{i=0}^{L-1}, \label{eq:kernel_matrix}
	\end{equation}
	\vspace{-0.3cm}
	\begin{equation}
		y = \mathbf{\overline{K}} \otimes x \label{eq:kernel_output},
	\end{equation}
	where \(\otimes\) denotes one-dimensional convolution. Mamba~\citep{gu2023mamba} implements this computation efficiently using either convolution or recurrence, and further enhances it with input-dependent parameterization, enabling scalable modeling for long sequences in applications such as language and video understanding.
	
	\subsection{Overall Framework}
	\subsubsection{DSAGL Architecture}
	\begin{figure}[h]
		\centering
		\includegraphics[width=\linewidth]{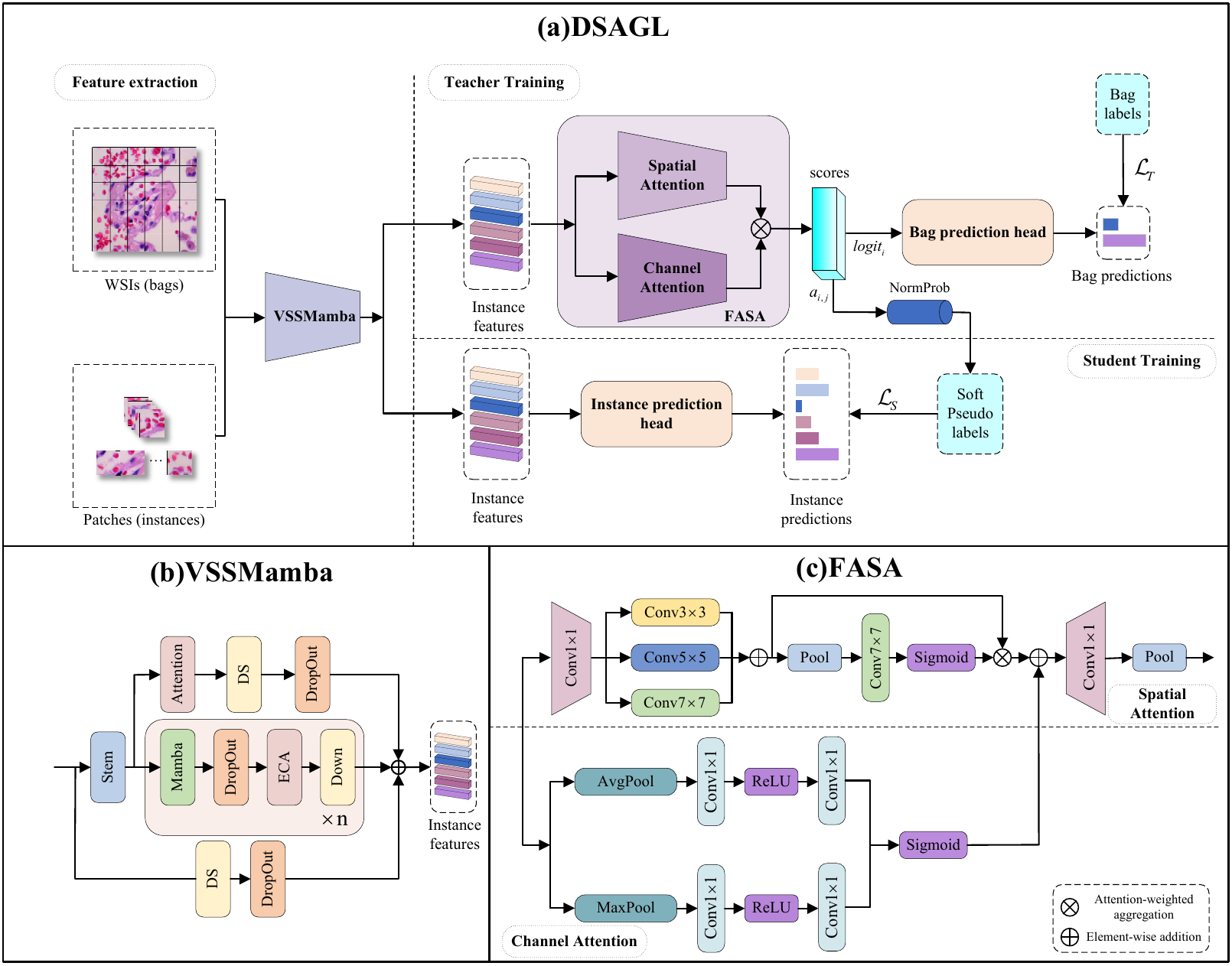}
		\caption{(a) The overall architecture of the proposed Dual-Stream Attention-Guided Learning Weakly Supervised Framework (DSAGL); (b) The proposed Vision-aware Skip-connected Selective Mamba encoder (VSSMamba); (c) The proposed Fusion-Attentive Scale-Aware module (FASA).}
		\label{fig:Overall network}
	\end{figure}

	To enable efficient and interpretable pathological image classification, we propose a Dual-Stream Attention-Guided Learning framework (DSAGL) for weakly supervised learning. As illustrated in Fig.~\ref{fig:Overall network}, the DSAGL architecture comprises two main components: a feature extraction module and a dual-stream network consisting of a teacher \rev{stream} and a student \rev{stream}. Specifically, we first adopt the Vision-aware Skip-connected Selective Mamba encoder (VSSMamba) to extract initial instance-level features from the input patches. These features are then independently processed by the teacher and student \rev{streams}. In the teacher \rev{stream}, we introduce the Fusion-Attentive Scale-Aware (FASA) module to generate bag-level predictions and compute attention scores, which are subsequently used to construct high-quality instance-level pseudo labels. The student \rev{stream} focuses on fine-grained instance-level classification, guided by the pseudo labels provided by the teacher. Through an alternating training strategy between the two \rev{streams}, DSAGL enables efficient collaborative feature learning and semantic alignment under weak supervision.
	
	\subsubsection{Alternating Training Strategy}
	
	\rev{To enhance collaborative optimization between the teacher and student streams under weak supervision, DSAGL adopts an alternating training strategy. Unlike conventional joint or sequential training paradigms, this strategy alternately updates the two streams at fixed intervals, enabling progressive refinement of pseudo labels and feature representations through bidirectional collaboration.}
	
	\rev{During training, the teacher stream and the student stream are optimized in a decoupled yet cooperative manner. The teacher stream focuses on global bag-level prediction and attention estimation, while the student stream specializes in fine-grained instance-level discrimination using the pseudo labels generated by the teacher. This staged update pattern mitigates potential gradient interference between the two streams and contributes to more stable convergence.}
	
	\rev{In practice, the alternating optimization is conducted at the epoch level. At each training epoch, the teacher stream is first updated using bag-level supervision to generate attention maps and instance-level pseudo labels. The student stream is then optimized within the same epoch using the pseudo labels produced by the current teacher. In all experiments, the student optimization period is set to one epoch, i.e., the teacher and student are alternately updated at a fixed epoch-wise interval. Both streams are randomly initialized at the beginning of training, without additional pretraining or warm-up stages.}
	
	\rev{This alternating strategy offers several key advantages. First, it enables dynamic refinement of pseudo labels, which are progressively improved during training, thereby reducing the impact of early-stage noise. Second, it achieves clear task decoupling, allowing the teacher to focus on holistic feature aggregation while the student learns detailed discriminative representations. Third, the staggered updates prevent simultaneous gradient conflicts and improve overall training stability. Finally, this design is particularly effective in weakly supervised scenarios, where sparse or noisy labels can severely compromise learning performance.}

	\subsection{VSSMamba (Feature Extraction)}
	Although CNNs~\citep{lecun1989backpropagation} and Transformers~\citep{vaswani2017attention} have excelled in vision tasks, they remain limited in modeling long-range dependencies and incur high computational costs on large-scale WSIs.To address these limitations, inspired by Vision Mamba~\citep{vim}, we propose VSSMamba, a new encoder that extracts discriminative instance-level features with linear complexity.
	
	As illustrated in part (b) of Fig.~\ref{fig:Overall network}, the VSSMamba comprises a main processing stream and two auxiliary skip connections, enabling enhanced multi-scale and multi-level context fusion. Given an input instance image $p \in \mathbb{R}^{\texttt{B} \times \texttt{C} \times \texttt{H} \times \texttt{W}}$, where \( \texttt{B} \), \( \texttt{C} \), \( \texttt{H} \), and \( \texttt{W} \) denote the batch size, number of channels, height, and width respectively, we first extract initial features via a convolutional stem layer:
	\begin{equation}
		f_0 = \mathrm{Stem}(p).
	\end{equation}
	
	\rev{As illustrated in Fig.~3(b), the stem output $f_0$ serves as the shared input to the main Mamba stream and the auxiliary skip branches. The features are subsequently processed by a stack of $n$ Mamba layers, denoted as $\{\mathcal{M}_i\}_{i=1}^{n}$, along the main path. Each layer consists of a Mamba transformation, followed by Dropout, Efficient Channel Attention $\mathcal{E}_i$, and a downsampling operation $\mathcal{D}_i$. The number of stacked layers $n$ can be flexibly adjusted based on data scale and task complexity to balance representation depth and capacity. We denote the final output of the main path as $f_n$ and output $f_{i+1}$ of the $i$-th layer is computed as:}
	\begin{equation}
		f_{i+1} = \mathcal{D}_i\left( \mathcal{E}_i\left( \mathrm{Drop}_i\left( \mathcal{M}_{i}(f_i) \right) \right) \right), \quad i = 0, 1, \dots, n{-}1.
	\end{equation}
	
	\rev{This operation corresponds to the central processing stream in Fig.~3(b), where each Mamba block is followed by channel attention, dropout, and downsampling. To enhance both local detail and global semantics, VSSMamba incorporates two skip connections, $f_{skip1}$ and $f_{skip2}$. The first skip path starts from the stem output $f_0$, and passes through a combined attention module consisting of Squeeze-and-Excitation and Spatial Attention $\mathcal{A}$. The output is then downsampled and regularized via Dropout to yield locally enhanced features. The second skip path directly downscales the raw input $p$ to extract global semantic cues:}
	\begin{equation}
		f_{skip1} = \mathrm{Drop}\left( \mathcal{D}\left( \mathcal{A}(f_0) \right) \right),
	\end{equation}
	\vspace{-0.3cm}
	\begin{equation}
		f_{skip2} = \mathrm{Drop}\left( \mathcal{D}(p) \right).
	\end{equation}
	
	Finally, VSSMamba integrates the outputs from all three paths---$f_n$, $f_{skip1}$, and $f_{skip2}$ through a residual aggregation strategy, effectively blending deep semantic features, localized attention-enhanced cues, and global contextual signals. This fusion yields the final instance-level representation $x$, which captures both fine-grained structural detail and long-range contextual dependencies crucial for accurate instance discrimination in WSI:
	\begin{equation}
		x = f_n + f_{skip1} + f_{skip2}.
	\end{equation}
	
	In summary, VSSMamba efficiently captures both global dependencies and local details through its hierarchical design with linear-time complexity. By combining the long-range modeling ability of Mamba layers with complementary skip-enhanced feature streams, it provides robust and expressive instance-level representations. This encoder serves as the backbone of our dual-stream framework, enabling effective learning under weak supervision for complex WSI classification tasks.

	\subsection{\rev{Dual-Stream Network}}
	To jointly model instance-level and bag-level semantics under weak supervision, we propose a dual-stream architecture comprising a teacher and a student \rev{stream}. This design is motivated by the inherent ambiguity in instance labels within WSIs, where relying solely on bag-level supervision often leads to suboptimal instance discrimination. By decoupling the representation learning into two cooperative \rev{streams}, the teacher can provide global guidance, while the student focuses on fine-grained instance-level feature extraction.
	
	In this section, we first present FASA, the key attention module designed to enhance the model's ability to extract discriminative features from sparse and ambiguous regions. FASA integrates multi-scale convolution and dual attention mechanisms to address the challenge of lesion heterogeneity and low signal-to-noise ratio. We then elaborate on the functional roles and collaborative interaction between the teacher and student \rev{streams}, followed by a detailed formulation of the hybrid loss function. This loss is tailored to mitigate the noise and uncertainty introduced by pseudo labels, thus promoting stable optimization and semantic alignment between the two streams.
	
	\subsubsection{FASA}
	To address the challenge of limited supervision, we introduce FASA, a fusion-attentive scale-aware module that enhances feature discriminability by integrating multi-scale convolutional fusion with both channel and spatial attention mechanisms. It performs attention-weighted global pooling to aggregate instance-level features, thereby facilitating discriminative learning in a weakly supervised setting.
	
	\rev{As illustrated in part (c) of Fig.~\ref{fig:Overall network}, FASA improves the discriminability of deep features by integrating spatial and channel attention into a multi-scale fusion framework. Given an input feature map \(x \in \mathbb{R}^{\texttt{B} \times \texttt{C} \times \texttt{H} \times \texttt{W}}\) extracted by the VSSMamba encoder, FASA captures local contextual information at multiple receptive fields through parallel convolutions with kernel sizes of \(3 \times 3\), \(5 \times 5\), and \(7 \times 7\), where \(\mathcal{C}_{k \times k}\) denotes a convolution operation of size \(k \times k\). The resulting multi-scale features are subsequently concatenated and aggregated along the channel dimension, yielding a unified representation \(h_{fused}\) that preserves complementary information across different spatial scales. Here, the summation in Eq.~(13) serves as a concise abstraction of this multi-scale feature fusion process, rather than a strict element-wise addition.}
	
	\rev{Next, channel and spatial attention mechanisms are independently applied to \(h_{fused}\), producing two complementary attention-enhanced feature maps: \(f_{channel}\) and \(f_{spatial}\). Here, \(A_t(\cdot)\) denotes a lightweight attention mapping function that generates a normalized channel-wise or spatial-wise attention map via a sigmoid activation, and the attention operation is implemented through element-wise feature reweighting. The spatial attention branch adopts a pooling-based fusion strategy, where both average pooling and max pooling are performed along the channel dimension to capture complementary spatial responses. These two attention-enhanced maps are then combined to obtain the final attentive representation \(f_{out}\), which integrates both modality-specific importance cues. During instance aggregation, \(f_{out}\) is subjected to global pooling to derive instance-wise attention weights \(a_{i,j}\) and the corresponding bag-level prediction score \(\text{logit}_i\). These outputs jointly serve as weak supervision signals for pseudo-label generation and bag-level classification.}
	
	%The detailed flow of feature computation is illustrated in the pseudocode shown in Algorithm~\ref{alg:fasa}.
	%\begin{comment}

	The flow of feature computation is formulated as follows:
	\begin{equation}
		h_{fused} = \sum_{k \in \{3, 5, 7\}} \mathcal{C}_{k \times k}(x), \quad k \in \{3, 5, 7\},
	\end{equation}
	\vspace{-0.3cm}
	\begin{equation}
		f_{t} = h_{fused} \times \mathcal{A}_t(h_{fused}), \quad t \in \{channel, spatial\},
	\end{equation}
	\vspace{-0.3cm}
	\begin{equation}
		f_{out} = f_{channel} + f_{spatial},
	\end{equation}
	\vspace{-0.3cm}
	\begin{equation}
		a_{i,j}, \, logit_i = \text{FASA}(f_{out}).
	\end{equation}
	
	By unifying multi-scale feature extraction with spatial and channel-wise attention, FASA provides a powerful mechanism for emphasizing diagnostically relevant regions under weak supervision. Its design allows the network to adaptively aggregate informative patterns across varying spatial resolutions, thereby reinforcing instance-level discriminability and improving the quality of attention-driven pseudo labels essential for downstream tasks.
	
	%\end{comment}
	\begin{comment}
		\definecolor{keywordcolor}{RGB}{0, 120, 0} % 关键词颜色
		\definecolor{commentcolor}{RGB}{128, 128, 128} % 注释颜色
		\definecolor{stringcolor}{RGB}{0, 0, 0} % 变量和操作符颜色
		\begin{algorithm}[H]
			\caption{FASA Block Process}
			\label{alg:fasa}
			\textbf{Input:} Feature map $x : \textcolor{keywordcolor}{\mathtt{(B, C, H, W)}}$ \\
			\textbf{Output:} Attention weights $a_{i,j} : \textcolor{keywordcolor}{\mathtt{(B, N)}}$, Prediction score $logit_i : \textcolor{keywordcolor}{\mathtt{(B, 1)}}$
			\begin{algorithmic}[1]
				\State $\textcolor{commentcolor}{\text{/* compute fused multi-scale features $h_{fused}$ */}}$
				\State $h_{fused} \gets \sum_{k \in \{3,5,7\}} C_{k \times k}(x)$
				
				\State $\textcolor{commentcolor}{\text{/* apply attention modules */}}$
				\For{$t \in \{channel, spatial\}$}
				\State $f_t \gets h_{fused} \bm{\odot} \mathcal{A}_t(h_{fused})$
				\EndFor
				
				\State $\textcolor{commentcolor}{\text{/* aggregate attention-modulated features */}}$
				\State $f_{out} \gets f_{channel} + f_{spatial}$
				
				\State $\textcolor{commentcolor}{\text{/* compute instance-level weights and prediction score */}}$
				\State $a_{i,j}, \textit{logit}_i \gets \text{FASA}(f_{out})$
				
				\State \Return $a_{i,j}, \textit{logit}_i$
			\end{algorithmic}
		\end{algorithm}
	\end{comment}
	\subsubsection{\rev{Teacher Stream}}
	The teacher \rev{stream} serves as a bag-level classifier that provides both global predictions and instance-wise guidance under weak supervision. This design enables the decoupling of coarse-grained supervision from fine-grained feature learning, allowing the teacher to learn stable bag-level semantics while generating informative cues for the student \rev{stream}.
	
	Upon receiving instance-level feature vectors, the teacher \rev{stream} employs FASA to compute attention scores for each instance and predicts the bag label via a classification head. The attention mechanism allows the teacher to focus on diagnostically relevant patches and suppress irrelevant or noisy regions, which is particularly important in WSIs where discriminative signals are sparse and spatially scattered.
	
	Specifically, for each instance \(\left\{ p_{i,j}, \, j = 1, 2, \dots, n_i \right\}\) in a bag \(B_i\), the corresponding feature vector \(x_{i,j}\) is passed into FASA to obtain its attention score \(a_{i,j}\) and the aggregated bag-level logit \(logit_i\). The attention scores \(\{a_{i,j}\}\) are propagated to the student \rev{stream} as soft supervisory signals, while the logit is further processed by the classification head \(P_t\) to generate the predicted bag label \(\hat{y}_i\). The bag prediction is supervised using the standard binary cross-entropy loss \(\mathcal{L}_{teacher}\), which is robust and widely adopted for binary weakly labeled classification tasks. The computation proceeds as follows:
	%\vspace{-0.1cm}
	\begin{equation}
		a_{i,j}, \, logit_i = \text{FASA}\left( x_{i,j} \mid x_{i,1}, x_{i,2}, \dots, x_{i,n_i} \right),
	\end{equation}
	\vspace{-0.3cm}
	\begin{equation}
		\hat{y}_i = P_t(logit_i),
	\end{equation}
	\vspace{-0.3cm}
	\begin{equation}
		\mathcal{L}_{T} = \mathcal{L}_{teacher}(y_i, \hat{y}_i).
	\end{equation}
	
	By decoupling semantic abstraction and region selection, the teacher \rev{stream} delivers reliable bag-level predictions while producing spatially-aware attention signals. These outputs play a pivotal role in guiding the student \rev{stream}, particularly under noisy and weakly annotated scenarios common in WSI tasks.
	
	\subsubsection{\rev{Student Stream}}
	
	The student \rev{stream} is designed to perform fine-grained instance-level classification under weak supervision. It receives instance feature vectors and predicts the class probability for each individual instance. While sharing the same feature extractor as the teacher \rev{stream}, the student maintains its own classification head \(P_s\), enabling independent optimization of instance-level representations.
	
	To guide the student in identifying diagnostically relevant regions, the attention score \(a_{i,j}\) provided by the teacher is passed through a normalization module, NormProb, which converts it into a probabilistic pseudo label \(\hat{z}_{i,j}\). This process stabilizes the supervision signal and helps mitigate early-stage noise in the attention distribution. The training of the student \rev{stream} is supervised using a hybrid loss function \(\mathcal{L}_{Hybrid}\), which combines cross-entropy loss~\citep{mao2023cross} with Kullback-Leibler divergence~\citep{kullback1951information} in a weighted formulation. This design allows the \rev{stream} to benefit from both hard pseudo labels and soft distribution alignment. The resulting student loss is denoted by \(\mathcal{L}_S\), and is computed as follows:

	\begin{equation}
		\hat{z}_{i,j} = \text{NormProb}(a_{i,j}),
	\end{equation}
	\begin{equation}
		p_{\hat{z}_{i,j}} = P_s(x_{i,j}),
	\end{equation}
	\begin{equation}
		\mathcal{L}_{S} = \mathcal{L}_{Hybrid}(\hat{z}_{i,j}, p_{\hat{z}_{i,j}}).
	\end{equation}

	By leveraging attention-informed pseudo labels and a tailored hybrid loss, the student \rev{stream} effectively refines instance-level predictions under weak supervision. Its design enables focused learning on diagnostically salient regions and complements the teacher \rev{stream} by reinforcing semantic consistency at a finer granularity.
	
	\subsubsection{Hybrid Loss Function}
	
	Under weak supervision, instance-level labels are unavailable, and pseudo labels generated by the teacher \rev{stream} are inherently noisy. Relying solely on cross-entropy (CE) loss can lead to overfitting and unstable optimization, especially in early training stages. To address this, we propose a hybrid loss function $\mathcal{L}_{Hybrid}$ that combines weighted hard-label supervision with soft-label regularization, balancing discriminative learning and robustness.
	
	The student \rev{stream} is supervised by two complementary objectives: a weighted cross-entropy loss $\mathcal{L}_{CE, student}$ that promotes confident classification based on hard pseudo labels, and a Kullback-Leibler (KL) divergence loss $\mathcal{L}_{KL, student}$ that aligns the student's output distribution with the teacher's soft predictions. This dual supervision enables sharper decision boundaries while preserving probabilistic consistency. To mitigate instance-level class imbalance, a weight factor $w_n$ is applied to negative samples in the CE loss.
	
	Formally, the teacher \rev{stream} is optimized with a bag-level cross-entropy loss:
	\begin{equation}
		\mathcal{L}_{teacher} = -\left[ y_i \log(\hat{y}_i + \epsilon) + (1 - y_i) \log(1 -\hat{y}_i + \epsilon) \right],
	\end{equation}
	where $y_i$ denotes the ground-truth bag label, $\hat{y}_i$ is the predicted bag-level probability, and $\epsilon$ is a small constant for numerical stability.
	%\vspace{-0.1cm}
	\begin{equation}
		\begin{split}
			\mathcal{L}_{CE, student} =\; & -\frac{1}{N} \sum_{i,j} \Big[
			w_n (1 - \hat{z}_{i,j}) \log(p_{i,j,0} + \epsilon) \\
			& \qquad +\, (1 - w_n) \hat{z}_{i,j} \log(p_{i,j,1} + \epsilon)
			\Big],
		\end{split}
	\end{equation}
	where $\hat{z}_{i,j} \in \{0,1\}$ is the hard pseudo label for instance $(i,j)$, $p_{i,j,c}$ is the student's predicted probability for class $c \in \{0,1\}$, and $w_n \in [0,1]$ controls the relative importance of negative instances.
	
	To leverage soft supervision, the KL divergence loss is introduced as:
	\begin{equation}
		\mathcal{L}_{KL, student} = T^2 \cdot \frac{1}{N} \sum_{i,j} \sum_{c \in \{0, 1\}} q_{i,j,c} \log \left( \frac{q_{i,j,c}}{p_{i,j,c}} \right),
	\end{equation}
	where $q_{i,j,c}$ and $p_{i,j,c}$ are the predicted class probabilities from the student and teacher, respectively. The temperature parameter $T > 1$ is used to soften the distributions, and the KL loss is scaled by $T^2$ to match the magnitude of CE gradients.
	
	The overall student loss is expressed as:
	\begin{equation}
		\mathcal{L}_{Hybrid} = \alpha \mathcal{L}_{CE, student} + (1 - \alpha) \mathcal{L}_{KL, student},
	\end{equation}
	where $\alpha \in [0,1]$ controls the trade-off between hard and soft supervision. The hybrid loss effectively stabilizes learning and improves generalization, particularly under noisy pseudo labels and imbalanced patch distributions.
	\rev{\subsection{Notation Summary}}
	\rev{To improve readability and ensure consistency between the mathematical formulation, network architecture, and structural diagrams, we summarize the main symbols used throughout the manuscript in Table~\ref{tab:notation_summary}. This table provides a unified reference for input representations, encoder components, teacher--student interactions, and optimization-related variables, and explicitly aligns key symbols with their corresponding modules illustrated in Fig.~\ref{fig:Overall network}.}
	
	\begin{table}[t]
		\centering
		\caption{\rev{Summary of mathematical symbols used in the manuscript.}}
		\label{tab:notation_summary}
		\setlength{\tabcolsep}{6pt}
		\begin{tabular}{c c p{8.0cm}}
			\toprule
			\textbf{Symbol} & \textbf{Eq./Fig.} & \textbf{Description} \\
			\midrule
			$p$ & Eq.~(8) & Input instance (image patch) \\
			$f_0$ & Eq.~(8), Fig.~3(b) & Feature map extracted by the stem layer \\
			$f_i$ & Eq.~(9), Fig.~3(b) & Intermediate feature after the $i$-th Mamba layer \\
			$f_n$ & Eq.~(9), Fig.~3(b) & Final output feature of the stacked Mamba layers \\
			$f_{skip1}$ & Eq.~(10), Fig.~3(b) & Skip feature derived from stem output \\
			$f_{skip2}$ & Eq.~(11), Fig.~3(b) & Skip feature derived from the raw input \\
			$x$ & Eq.~(12), Fig.~3(b) & Final instance-level representation after feature fusion \\
			$\mathcal{M}_i(\cdot)$ & Eq.~(9) & Mamba-based transformation in the main stream \\
			$\mathcal{E}_i(\cdot)$ & Eq.~(9) & Channel-wise attention operation \\
			$\mathcal{D}_i(\cdot)$ & Eq.~(9) & Downsampling operation \\
			$\mathrm{Drop}_i(\cdot)$ & Eq.~(9) & Dropout operation applied at the $i$-th layer \\
			$h_{\text{fused}}$ & Eq.~(13) & Fused multi-scale feature representation \\
			$C_{k\times k}(\cdot)$ & Eq.~(13) & Convolution operation with kernel size $k\times k$ \\
			$A_t(\cdot)$ & Eq.~(14) & Attention function for $t \in \{\text{channel}, \text{spatial}\}$ \\
			$f_{\text{channel}}$ & Eq.~(15) & Channel-attention-enhanced feature map \\
			$f_{\text{spatial}}$ & Eq.~(15) & Spatial-attention-enhanced feature map \\
			$f_{\text{out}}$ & Eq.~(15),(16) & Final attention-enhanced feature representation \\
			$a_{i,j}$ & Fig.~3(a), Eq.~(16),(17) & Attention score of the $j$-th instance in the $i$-th bag \\
			$\mathrm{logit}_i$ & Eq.~(16)--(18) & Aggregated bag-level logit for the $i$-th bag \\
			$x_{i,j}$ & Eq.~(17),(21) & Feature vector of the $j$-th instance in bag $i$ \\
			$\hat{y}_i$ & Eq.~(18),(19) & Predicted bag-level label \\
			$P_t(\cdot)$ & Eq.~(18) & Teacher prediction function \\
			$P_s(\cdot)$ & Eq.~(21) & Student prediction function \\
			$\tilde{y}_{i,j}$ & Fig.~3(a) & Soft pseudo label generated by the teacher stream \\
			$\hat{z}_{i,j}$ & Eq.~(20),(22),(24) & Normalized pseudo label for instance $(i,j)$ \\
			$p_{i,j,c}$ & Eq.~(24),(25) & Student-predicted probability for class $c$ \\
			$q_{i,j,c}$ & Eq.~(25) & Teacher-predicted probability for class $c$ \\
			$\mathcal{L}_T$ & Fig.~3(a), Eq.~(19),(23) & Teacher loss \\
			$\mathcal{L}_S$ & Fig.~3(a),(22) & Student loss \\
			$\mathcal{L}_{CE,\text{student}}$ & Eq.~(24) & Weighted cross-entropy loss for the student stream \\
			$\mathcal{L}_{KL,\text{student}}$ & Eq.~(25) & KL divergence loss for soft supervision \\
			$\mathcal{L}_{Hybrid}$ & Eq.~(22),(26) & Hybrid loss combining CE and KL terms \\
			$\alpha$ & Eq.~(26) & Trade-off weight between hard and soft supervision \\
			$T$ & Eq.~(25) & Temperature parameter for soft distribution smoothing \\
			$w_n$ & Eq.~(24) & Weight for negative instances \\
			$N$ & Eq.~(24),(25) & Total number of instances \\
			$c$ & Eq.~(24),(25) & Class index \\
			$\epsilon$ & Eq.~(23),(24) & Small constant for numerical stability \\
			$\oplus$ & Fig.~3(b),(c) & Element-wise addition \\
			$\otimes$ & Fig.~3(a),(c) & Attention-weighted aggregation \\
			\bottomrule
		\end{tabular}
	\end{table}

	\section{Experiment and Results }
	\subsection{Datasets}
\begin{figure}[h]
		\centering
		\includegraphics[width=0.6\linewidth]{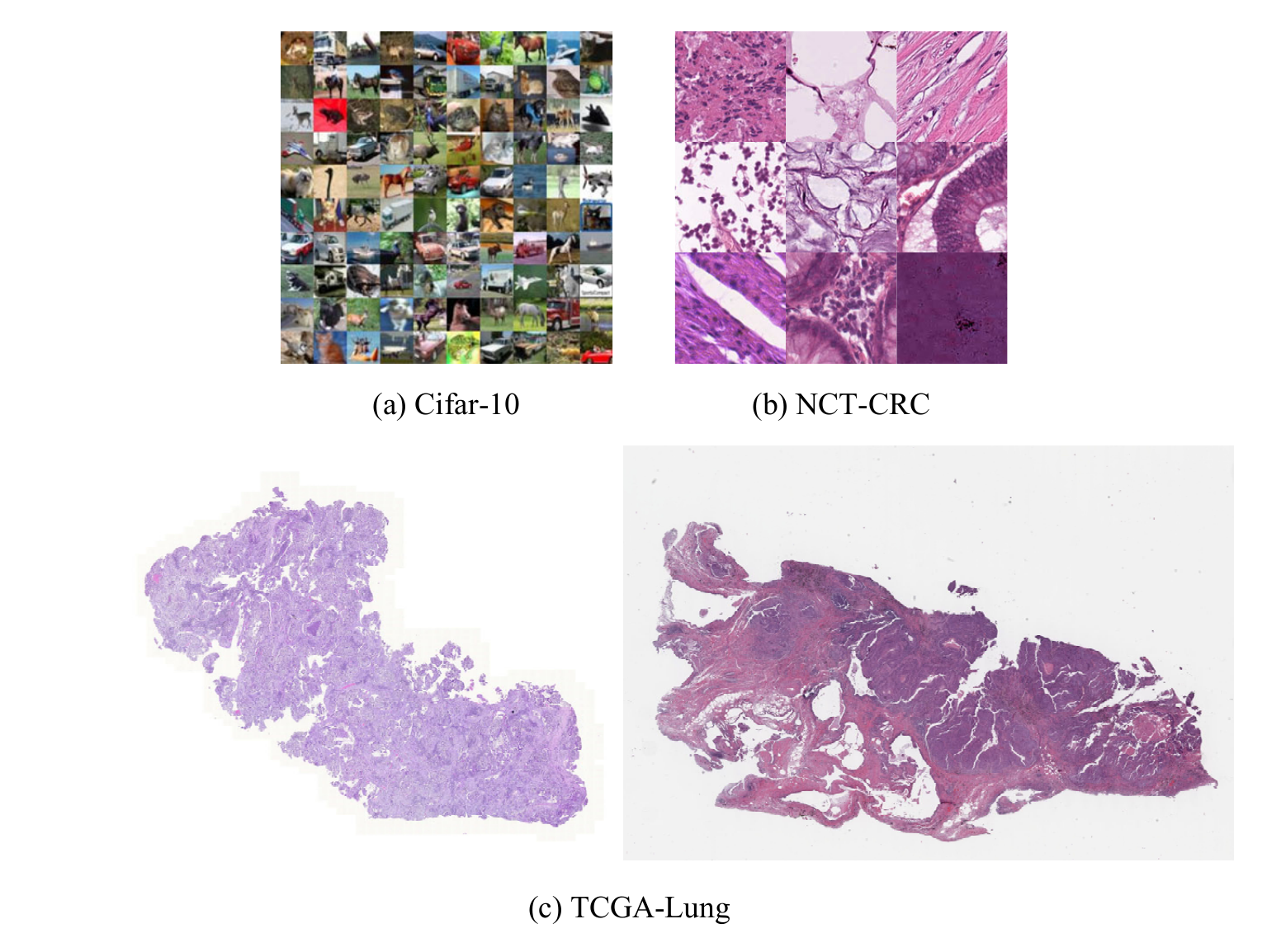}
		\caption{Examples of the three datasets used in our experiments. (a) CIFAR-10, consisting of 10 categories of natural images; (b) NCT-CRC, consisting of 9 categories of colorectal histopathology image patches; (c) TCGA-Lung, consisting of whole-slide images from two lung cancer subtypes.}
		\label{fig:dataset}
	\end{figure}

	To comprehensively evaluate the scalability and generalization of the proposed DSAGL framework under different image resolutions and weak supervision settings, we conducted experiments on three datasets of progressively increasing complexity, as illustrated in Fig.~\ref{fig:dataset}. Specifically, these datasets were selected to simulate a spectrum of instance resolutions ranging from low-level synthetic images to high-resolution WSIs, enabling a more systematic assessment of model robustness and adaptability.
	
	\rev{It is important to emphasize that CIFAR-10~\citep{krizhevsky2009learning} is not introduced as a surrogate or approximate simulation of real pathological images, but rather as a highly controllable synthetic benchmark for weakly supervised learning. Owing to its low instance resolution, limited spatial context, and the ability to systematically adjust the level of weak-supervision noise, CIFAR-10 is particularly well suited for analyzing optimization behavior and training stability under extreme conditions. In this setting, the dataset enables targeted investigation of how the dual-stream teacher–student architecture, attention-guided pseudo supervision, and alternating optimization strategy contribute to alleviating instance-level label ambiguity, reducing early-stage pseudo-label noise, and stabilizing training dynamics. The key insights derived from CIFAR-10 experiments do not pertain to pathological texture patterns or morphological structures; instead, they focus on optimization-level properties that are independent of specific visual semantics. As such, these insights are transferable across different image resolutions and domains.}
	
	NCT-CRC-HE-100K~\citep{kather2016multi} represents a medium-resolution histopathology dataset, providing a bridge between synthetic and real pathological data. It enables evaluation of DSAGL’s adaptability to morphological variations and inter-class heterogeneity in colorectal tissue images.
	
	Finally, the TCGA-Lung~\citep{li2021dual} dataset comprises gigapixel-level WSIs from lung adenocarcinoma (LUAD) and lung squamous cell carcinoma (LUSC) patients. It serves as a real-world benchmark for large-scale weakly supervised classification, verifying DSAGL’s capability to handle high-resolution pathological slides with limited annotations.
	
	Together, these three datasets provide a comprehensive and hierarchical evaluation setting, allowing DSAGL to be assessed across multiple image resolutions, levels of supervision, and domains from synthetic benchmarks to real-world WSIs.
	
	\subsubsection{CIFAR-10}
	CIFAR-10~\citep{krizhevsky2009learning} is a standard image classification dataset consisting of 60,000 color images of size $32\times32$ pixels, divided into 10 classes, with 6,000 images per class. Among them, 50,000 images are used for training and 10,000 for testing. The training set is organized into five batches of 10,000 images each, while the test set constitutes a separate batch. In weakly supervised learning tasks, CIFAR-10 is commonly utilized to simulate WSIs by randomly assembling images from different classes to construct synthetic datasets. In our setting, we simulate WSIs by randomly combining images from different classes within the dataset.
	
	\subsubsection{NCT-CRC}
	NCT-CRC-HE-100K~\citep{kather2016multi} is a real-world histopathology image dataset comprising 100,000 non-overlapping image patches extracted from 86 H\&E-stained human colorectal cancer and normal tissue slides. All images are of size $224\times224$ pixels and sampled at $20\times$ magnification. Each image is annotated with one of nine histological tissue classes: adipose (ADI), background (BACK), debris (DEB), lymphocytes (LYM), mucus (MUC), smooth muscle (MUS), normal colon mucosa (NORM), cancer-associated stroma (STR), and colorectal adenocarcinoma epithelium (TUM). Based on this dataset, we constructed multiple subsets with varying positive instance ratios by selecting images from different tissue classes.
	
	\subsubsection{TCGA-Lung}
	The TCGA (The Cancer Genome Atlas) lung cancer dataset~\citep{li2021dual} is an important publicly available resource for histopathological image analysis, encompassing two major subtypes of non-small cell lung cancer: LUAD and LUSC. The dataset comprises a total of 1,054 high-resolution WSIs stained with hematoxylin and eosin (H\&E), provided in the svs format and accessible through the data portal of the National Cancer Institute of the United States. At $20\times$ magnification, each WSI is partitioned into approximately 5.2 million image patches of size $224\times224$ pixels. Each WSI constitutes a bag in the multiple instance learning setting, containing a varying number of tumor and normal patches.
	
	\subsection{Experimental Details}
	\rev{In the experimental setup, we observe that the instance resolution in the CIFAR-10 dataset is relatively low, which makes it less suitable for the Mamba architecture designed for modeling long-range dependencies. Consequently, the number of Mamba layers along the main path (as illustrated in Fig.~\ref{fig:Overall network}) is set to zero, while the convolutional depth of the Stem module is increased to enhance feature extraction capability. In contrast, for the NCT-CRC and TCGA-Lung datasets, where the instance resolution is higher, the number of Mamba layers is set to 2 and 3, respectively.}
	
	\rev{To facilitate stable optimization and efficient parameter updates, we employ the Adam optimizer~\citep{kingma2014adam} for experiments on the NCT-CRC dataset, with a fixed learning rate of 0.0001, a weight decay of $1\times10^{-5}$, and a batch size of 4. For CIFAR-10 and TCGA-Lung, we adopt the SGD optimizer~\citep{robbins1951stochastic} with a learning rate of 0.001. Due to the large-scale nature of whole slide images in TCGA-Lung, the batch size is set to 1. In addition to GPU memory constraints, this choice is also statistically reasonable: each WSI contains tens of thousands of instances, such that a single bag already provides a rich empirical distribution of patches. Consequently, gradient estimation at the bag level remains stable even with a batch size of one, which is a common and well-established practice in WSI-based MIL frameworks. All models are trained for 1500 epochs. All experiments are conducted on a single NVIDIA Tesla V100 GPU equipped with 32~GB memory.}
	
	\rev{In the dual-stream optimization process, the total loss consists of the cross-entropy (CE) loss and the Kullback–Leibler (KL) divergence loss. To balance supervised classification and inter-stream consistency, different loss weights are applied. Empirically, the distillation temperature $T$ is fixed at 1.0 to maintain stable gradient scaling. The fusion weight $\alpha$ linearly increases from 0.9 to 1.0 as training progresses, emphasizing the KL loss in early epochs for pseudo-label stabilization and gradually strengthening the CE loss to enhance discriminative learning. In addition, to mitigate the impact of noisy negative instances in weakly supervised learning, the negative weighting factor in the student loss is set to $w_n = 0.10$. This dynamic weighting strategy effectively alleviates the influence of noisy pseudo labels and improves overall convergence stability.}
	
	\rev{For both instance-level and bag-level classification tasks, we adopt the area under the ROC curve (AUC)~\citep{hanley1982meaning} as the primary evaluation metric. Specifically, we report the AUC scores of DSAGL on both instance-level and bag-level tasks for the benchmark datasets. For CIFAR-10 and NCT-CRC, instance-level AUC is evaluated using the available ground-truth instance labels. These instance annotations are never used during training and are accessed only at evaluation time to objectively assess instance-level discriminability under weak supervision, following common practice in MIL benchmarking. All models are evaluated on the CIFAR-10 and NCT-CRC datasets under varying positive instance ratios. Specifically, CIFAR-10 is tested under ratios ranging from 1\% to 70\%, while NCT-CRC uses ratios from 10\% to 70\%. However, since the TCGA-Lung cancer dataset does not provide ground-truth instance-level labels, we only report the bag-level AUC for this dataset.}
		
	\rev{To assess the robustness of the experimental results, all experiments are repeated multiple times with different random seeds. Specifically, experiments on CIFAR-10 and NCT-CRC are conducted with five different random seeds, while experiments on the TCGA-Lung dataset are repeated three times due to its large computational cost. We report the mean and standard deviation (mean $\pm$ std) of the AUC scores across multiple runs. For CIFAR-10 and NCT-CRC, all baseline methods are reimplemented and trained under a unified experimental setting, including the same backbone architecture, data preprocessing, optimizer, and training schedule, to ensure fair comparison. For the TCGA-Lung dataset, due to the high computational cost of gigapixel WSIs, several baseline results are reported using the backbone configurations and training protocols specified in their original publications, which is consistent with common practice in WSI benchmarking. In addition, paired statistical significance tests are conducted to assess whether the performance improvements of DSAGL over competing methods are statistically reliable. Specifically, paired t-tests are applied to the AUC scores obtained from multiple runs with different random seeds. Improvements with a p-value smaller than 0.05 are considered statistically significant.}
	
	\subsection{Comparison Methods}
	
	\rev{To systematically evaluate the effectiveness of the proposed DSAGL framework, we compare it against a range of representative multiple instance learning (MIL) approaches. Based on their architectural characteristics, these methods can be categorized into three groups:}
	
	\rev{\textbf{Classical MIL Models.} Representative methods such as MI-Net~\citep{wang2018revisiting} and RNN-MIL~\citep{campanella2019clinical} adopt fixed pooling strategies (e.g., max or mean pooling) to aggregate instance features into bag-level representations. While computationally efficient, these methods lack mechanisms for modeling discriminative regions explicitly and often underperform in weakly supervised scenarios.}
	
	\rev{\textbf{Attention-based MIL Models.} Attention-MIL and its variants, including ABMIL and Gated-ABMIL~\citep{ilse2018attention}, introduce learnable attention modules to assign instance-specific weights, enabling flexible aggregation and improved interpretability. CLAM~\citep{lu2021data} further enhances localization by incorporating attention pruning to focus on diagnostically relevant patches.}
	
	\rev{\textbf{Transformer-based or Advanced MIL Models.} This category enhances MIL performance by integrating global context modeling and architectural refinements. TransMIL~\citep{shao2021transmil} leverages self-attention to capture inter-instance dependencies; IBMIL~\citep{lin2023interventional} introduces region-level interventions for improved instance discrimination; R$^2$T-MIL~\citep{tang2024feature} adopts token-level guidance to promote region-aware learning. Our proposed DSAGL also falls into this category, combining dual-stream optimization, attention-guided pseudo supervision, and efficient long-range modeling via the VSSMamba encoder.}
	\rev{\subsubsection{Mechanism-level Comparison with Related MIL Methods}}
	\rev{To more clearly elucidate the fundamental differences between the proposed DSAGL framework and existing multiple instance learning approaches, we conduct a mechanism-level comparison along several key dimensions, including supervision granularity, pseudo-label generation, teacher–student interaction, and training strategy, as summarized in Table~\ref{tab:mechanism_comparison}.}
	
	\rev{First, unlike classical dual-branch MIL methods such as DSMIL, which primarily rely on implicit top-instance selection to bridge bag-level and instance-level representations, DSAGL explicitly propagates bag-level supervision to the instance level through an attention-guided pseudo-labeling mechanism. By progressively refining instance-level supervision during training, DSAGL enables continuous correction of instance decision boundaries, thereby mitigating the severe label ambiguity inherent in weakly supervised settings.}
	
	\rev{Second, compared with token-level distillation approaches such as R2T-MIL, which treat the teacher model as a static source of supervision, DSAGL introduces an alternating optimization strategy that establishes a dynamic and bidirectional interaction between the bag-level teacher and the instance-level student. By alternately updating the two streams, DSAGL effectively alleviates gradient conflicts commonly observed in joint optimization schemes and improves the stability and reliability of the generated pseudo labels.}
	
	\rev{Furthermore, in contrast to knowledge distillation frameworks such as GLMKD that employ static distillation objectives or multi-teacher consensus, the core advantage of DSAGL lies not in introducing additional teacher branches, but in its training dynamics. The alternating training mechanism allows the bag-level semantic representation learned by the teacher to evolve synchronously with the improving discriminative capability of the student, thus avoiding the limitations imposed by fixed distillation targets.}
	
	\rev{Regarding long-range dependency modeling, Transformer-based MIL methods such as TransMIL leverage self-attention to capture global inter-instance relationships. While effective, their computational complexity scales quadratically with the number of instances, which can hinder scalability in large-scale whole slide image analysis. In contrast, DSAGL adopts a VSSMamba-based encoder with linear complexity, achieving efficient global context modeling while significantly improving inference efficiency and scalability.}
	
	\rev{Overall, the performance gains of DSAGL do not stem from a single architectural modification or backbone replacement, but rather from the synergistic integration of attention-guided instance-level pseudo supervision and alternating training dynamics. This mechanism-level distinction enables DSAGL to achieve a more balanced trade-off between discriminative performance and computational efficiency in weakly supervised whole slide image classification.}
	
	\begin{table}[t]
		\centering
		\caption{\rev{Mechanism-level comparison between DSAGL and representative MIL methods.}}
		\label{tab:mechanism_comparison}
		\scriptsize
		\setlength{\tabcolsep}{4pt}
		\begin{tabular}{lccccc}
			\toprule
			\textbf{Method}
			& \textbf{Supervision \& Pseudo-labeling}
			& \textbf{Teacher--Student Scheme}
			& \textbf{Training Strategy}
			& \textbf{Long-range Modeling}
			& \textbf{Complexity} \\
			\midrule
			DSMIL
			& Bag + top-instance (implicit)
			& Static dual-branch
			& Joint
			& --
			& Low \\
			
			R2T-MIL
			& Bag + token attention
			& Token-level distillation
			& Joint / Sequential
			& Self-attention
			& High \\
			
			GLMKD
			& Bag + region consensus
			& Multi-teacher KD
			& Joint
			& --
			& Medium \\
			
			TransMIL
			& Bag-level only
			& --
			& Joint
			& Self-attention
			& High \\
			
			\textbf{DSAGL (Ours)}
			& \textbf{Bag $\rightarrow$ instance (progressive)}
			& \textbf{Bidirectional alternating}
			& \textbf{Alternating}
			& \textbf{VSSMamba (linear)}
			& \textbf{Moderate} \\
			\bottomrule
		\end{tabular}
	\end{table}

	\subsection{Experimental Results}
	\subsubsection{Results on the CIFAR-10 Dataset}
	
\begin{figure}[h]
		\centering
		\begin{subfigure}[b]{0.45\textwidth} % 调整子图宽度
			\centering
			\includegraphics[width=1.0\textwidth]{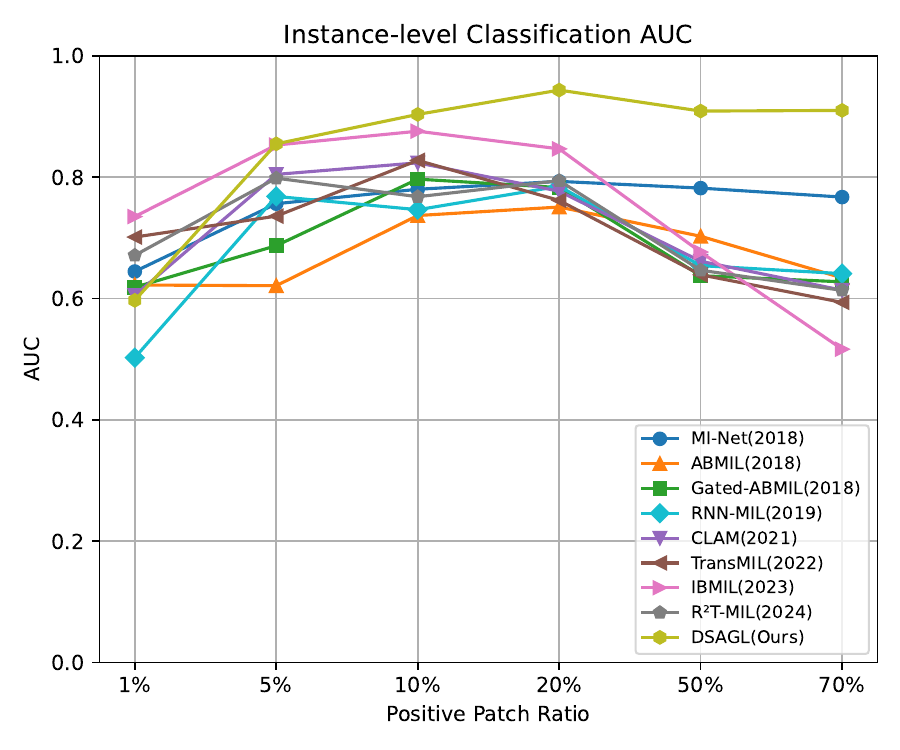} % 调整图片宽度
			\caption{Instance-level Classification AUC.}
			\label{fig:ins_auc}
		\end{subfigure}
		\hspace{15pt} % 控制两图之间的距离
		\begin{subfigure}[b]{0.45\textwidth} % 调整子图宽度
			\centering
			\includegraphics[width=1.0\textwidth]{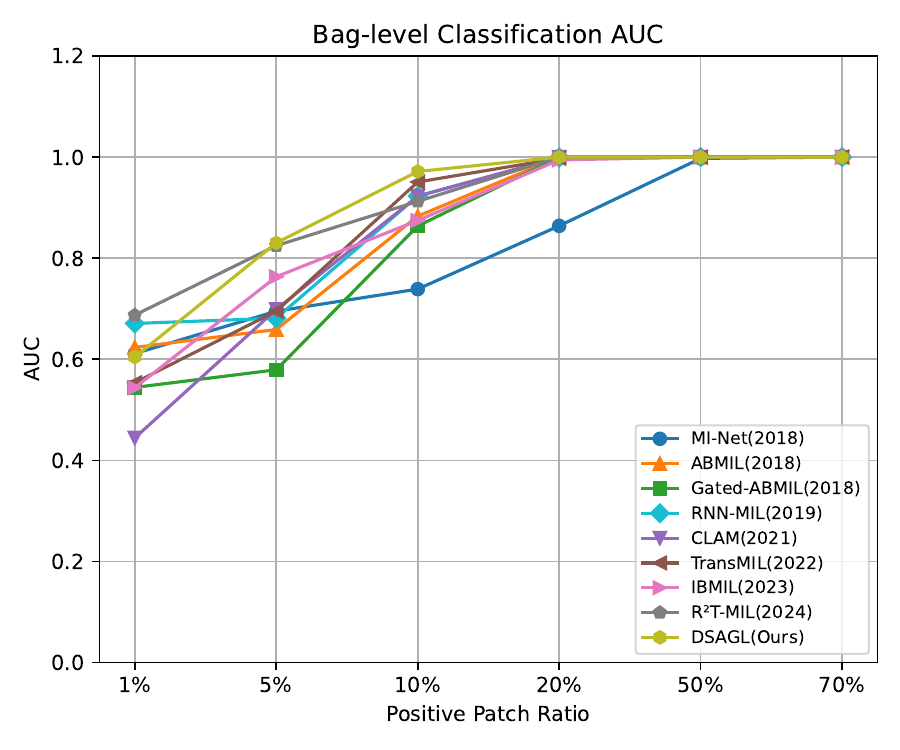} % 调整图片宽度
			\caption{Bag-level Classification AUC.}
			\label{fig:bag_auc}
		\end{subfigure}
		\caption{Comparison of AUC for Different Classification Levels on the CIFAR-10 dataset.}
		\label{fig:combined_auc}
	\end{figure}

	\rev{To evaluate the optimization stability and training behavior of the proposed framework under controllable weak supervision, we conduct experiments on the synthetic CIFAR-10 dataset with varying positive patch ratios ranging from 1\% to 70\%. As summarized in Table \ref{tab:cifar}, DSAGL achieves competitive performance at both the instance and bag levels across all settings. In particular, under medium-to-high positive ratios ($\geq$10\%), DSAGL consistently outperforms representative MIL baselines, including ABMIL, CLAM, TransMIL, and IBMIL, in terms of instance-level AUC. For example, at positive ratios of 10\% and 20\%, DSAGL attains instance-level AUCs of about 0.9019 and 0.9268, respectively, whereas CLAM and TransMIL remain below approximately 0.83 under the same conditions. At the bag level, when the positive ratio reaches 20\% or higher, DSAGL achieves near-saturated performance with an AUC of about 1.0000, indicating strong robustness to instance-level noise and class imbalance in this synthetic weakly supervised setting. Figure \ref{fig:combined_auc} visualizes the performance trends across different positive ratios.}
	
\begin{figure}[h]
		\centering
		\begin{subfigure}[b]{0.45\textwidth} % 调整子图宽度
			\centering
			\includegraphics[width=1.0\textwidth]{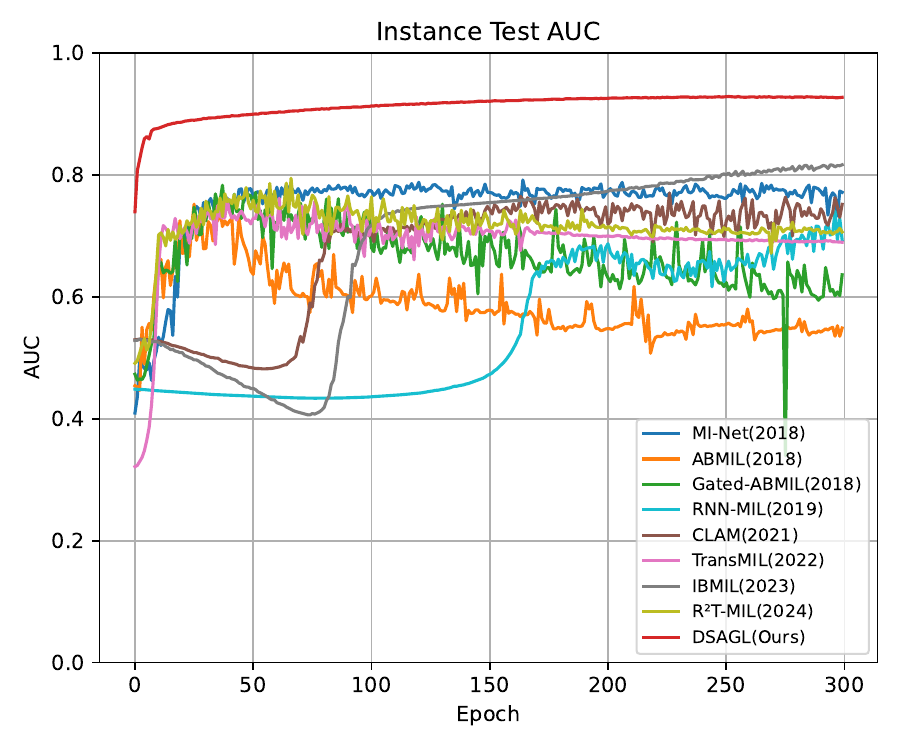} % 调整图片宽度
			\caption{Instance-level Classification AUC.}
			\label{fig:ins_auc}
		\end{subfigure}
		\hspace{15pt} % 控制两图之间的距离
		\begin{subfigure}[b]{0.45\textwidth} % 调整子图宽度
			\centering
			\includegraphics[width=1.0\textwidth]{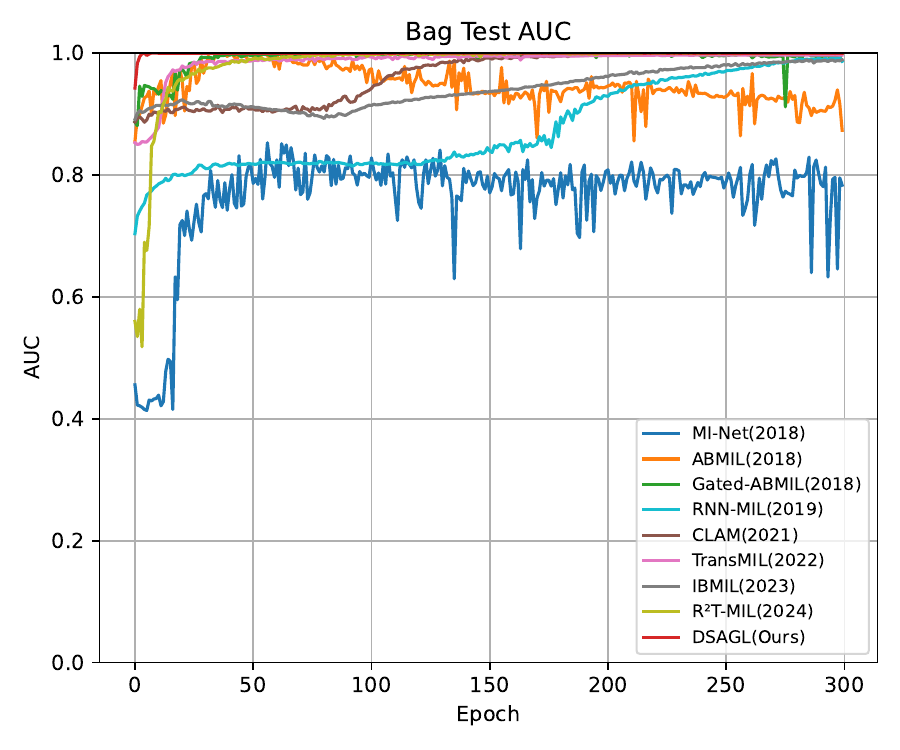} % 调整图片宽度
			\caption{Bag-level Classification AUC.}
			\label{fig:bag_auc}
		\end{subfigure}
		\caption{Test AUC curves on the CIFAR-10 dataset. }
		\label{fig:combined_auc2}
	\end{figure}

	\rev{Furthermore, we analyze the training dynamics under a representative setting with a 20\% positive patch ratio by plotting the test AUC curves over 300 training epochs (Figure \ref{fig:combined_auc2}). To ensure fair comparison, all baseline methods are reimplemented under identical conditions, including the same feature extractor, optimizer, and training schedule. As observed, DSAGL converges more rapidly and exhibits a smoother performance evolution during training. Specifically, its instance-level AUC exceeds about 0.90 within approximately 100 epochs and remains stable thereafter. In contrast, classical MIL methods such as MI-Net and ABMIL converge more slowly, with instance-level AUCs remaining below about 0.80 for a substantial portion of training, while more complex models such as CLAM and TransMIL show noticeable fluctuations or early saturation in later epochs.}
	
	\begin{table}[h]
		\centering
		\scriptsize
		\caption{Results on the CIFAR-10 dataset.}
		\label{tab:cifar}
		
		\begin{subtable}[t]{\textwidth}
			\centering
			\caption{Instance-level classification AUC.}
			\setlength{\tabcolsep}{1pt}
			\begin{tabular}{lcccccc}
				\toprule
				\textbf{Positive patch ratio} & 1\% & 5\% & 10\% & 20\% & 50\% & 70\% \\
				\midrule
				\multicolumn{7}{l}{\textbf{\textit{Classical MIL Models}}} \\
				MI-Net(2018)~\citep{wang2018revisiting}
				& 0.6429 \ensuremath{\pm} 0.0021 & 0.7546 \ensuremath{\pm} 0.0019 & 0.7786 \ensuremath{\pm} 0.0018
				& 0.7914 \ensuremath{\pm} 0.0017 & 0.7803 \ensuremath{\pm} 0.0016 & 0.7655 \ensuremath{\pm} 0.0019 \\
				RNN-MIL(2019)~\citep{campanella2019clinical}
				& 0.5010 \ensuremath{\pm} 0.0025 & 0.7668 \ensuremath{\pm} 0.0022 & 0.7448 \ensuremath{\pm} 0.0021
				& 0.7833 \ensuremath{\pm} 0.0019 & 0.6526 \ensuremath{\pm} 0.0020 & 0.6394 \ensuremath{\pm} 0.0022 \\
				\midrule
				\multicolumn{7}{l}{\textbf{\textit{Attention-based MIL Models}}} \\
				ABMIL(2018)~\citep{ilse2018attention}
				& 0.6208 \ensuremath{\pm} 0.0023 & 0.6196 \ensuremath{\pm} 0.0021 & 0.7352 \ensuremath{\pm} 0.0019
				& 0.7534 \ensuremath{\pm} 0.0018 & 0.7010 \ensuremath{\pm} 0.0020 & 0.6324 \ensuremath{\pm} 0.0022 \\
				Gated-ABMIL(2018)~\citep{ilse2018attention}
				& 0.6170 \ensuremath{\pm} 0.0024 & 0.6856 \ensuremath{\pm} 0.0020 & 0.7951 \ensuremath{\pm} 0.0018
				& 0.7812 \ensuremath{\pm} 0.0019 & 0.6360 \ensuremath{\pm} 0.0021 & 0.6259 \ensuremath{\pm} 0.0023 \\
				CLAM(2021)~\citep{lu2021data}
				& 0.6037 \ensuremath{\pm} 0.0026 & 0.8030 \ensuremath{\pm} 0.0018 & 0.8221 \ensuremath{\pm} 0.0017
				& 0.7752 \ensuremath{\pm} 0.0019 & 0.6598 \ensuremath{\pm} 0.0020 & 0.6121 \ensuremath{\pm} 0.0022 \\
				\midrule
				\multicolumn{7}{l}{\textbf{\textit{Transformer-based or Advanced MIL Models}}} \\
				TransMIL(2021)~\citep{shao2021transmil}
				& 0.6997 \ensuremath{\pm} 0.0021 & 0.7342 \ensuremath{\pm} 0.0020 & 0.8261 \ensuremath{\pm} 0.0018
				& 0.7601 \ensuremath{\pm} 0.0019 & 0.6375 \ensuremath{\pm} 0.0021 & 0.5921 \ensuremath{\pm} 0.0023 \\
				IBMIL(2023)~\citep{lin2023interventional}
				& \textbf{0.7336 \ensuremath{\pm} 0.0019} & 0.8514 \ensuremath{\pm} 0.0017 & 0.8742 \ensuremath{\pm} 0.0016
				& 0.8451 \ensuremath{\pm} 0.0017 & 0.6750 \ensuremath{\pm} 0.0020 & 0.5149 \ensuremath{\pm} 0.0024 \\
				R\textsuperscript{2}T-MIL(2024)~\citep{tang2024feature}
				& 0.6697 \ensuremath{\pm} 0.0022 & 0.7969 \ensuremath{\pm} 0.0019 & 0.7663 \ensuremath{\pm} 0.0020
				& 0.7927 \ensuremath{\pm} 0.0018 & 0.6451 \ensuremath{\pm} 0.0021 & 0.6118 \ensuremath{\pm} 0.0022 \\
				DSAGL(Ours)
				& 0.5954 \ensuremath{\pm} 0.0023
				& \textcolor{red}{\textbf{0.8535 \ensuremath{\pm} 0.0015}}
				& \textcolor{red}{\textbf{0.9019 \ensuremath{\pm} 0.0014}}
				& \textcolor{red}{\textbf{0.9268 \ensuremath{\pm} 0.0013}}
				& \textcolor{red}{\textbf{0.9074 \ensuremath{\pm} 0.0014}}
				& \textcolor{red}{\textbf{0.9086 \ensuremath{\pm} 0.0015}} \\
				\bottomrule
			\end{tabular}
		\end{subtable}
		
		\vspace{6pt} % 控制两个子表之间的竖直间距\\
		
		\begin{subtable}[t]{\textwidth}
			\centering
			\caption{Bag-level classification AUC.}
			\setlength{\tabcolsep}{1pt}
			\begin{tabular}{lcccccc}
				\toprule
				\textbf{Positive patch ratio} & 1\% & 5\% & 10\% & 20\% & 50\% & 70\% \\
				\midrule
				\multicolumn{7}{l}{\textbf{\textit{Classical MIL Models}}} \\
				MI-Net(2018)~\citep{wang2018revisiting}
				& 0.6097 \ensuremath{\pm} 0.0022 & 0.6935 \ensuremath{\pm} 0.0020 & 0.7370 \ensuremath{\pm} 0.0018
				& 0.8621 \ensuremath{\pm} 0.0015 & 0.9057 \ensuremath{\pm} 0.0013 & \textbf{1.0000 \ensuremath{\pm} 0.0000} \\
				RNN-MIL(2019)~\citep{campanella2019clinical}
				& 0.6691 \ensuremath{\pm} 0.0021 & 0.6795 \ensuremath{\pm} 0.0020 & 0.9212 \ensuremath{\pm} 0.0016
				& 0.9974 \ensuremath{\pm} 0.0006 & \textbf{1.0000 \ensuremath{\pm} 0.0000} & \textbf{1.0000 \ensuremath{\pm} 0.0000} \\
				\midrule
				\multicolumn{7}{l}{\textbf{\textit{Attention-based MIL Models}}} \\
				ABMIL(2018)~\citep{ilse2018attention}
				& 0.6217 \ensuremath{\pm} 0.0023 & 0.6571 \ensuremath{\pm} 0.0021 & 0.8814 \ensuremath{\pm} 0.0017
				& \textbf{1.0000 \ensuremath{\pm} 0.0000} & \textbf{1.0000 \ensuremath{\pm} 0.0000} & \textbf{1.0000 \ensuremath{\pm} 0.0000} \\
				Gated-ABMIL(2018)~\citep{ilse2018attention}
				& 0.5426 \ensuremath{\pm} 0.0024 & 0.5775 \ensuremath{\pm} 0.0022 & 0.8619 \ensuremath{\pm} 0.0018
				& \textbf{1.0000 \ensuremath{\pm} 0.0000} & \textbf{1.0000 \ensuremath{\pm} 0.0000} & \textbf{1.0000 \ensuremath{\pm} 0.0000} \\
				CLAM(2021)~\citep{lu2021data}
				& 0.4428 \ensuremath{\pm} 0.0026 & 0.6968 \ensuremath{\pm} 0.0021 & 0.9209 \ensuremath{\pm} 0.0017
				& \textbf{1.0000 \ensuremath{\pm} 0.0000} & \textbf{1.0000 \ensuremath{\pm} 0.0000} & \textbf{1.0000 \ensuremath{\pm} 0.0000} \\
				\midrule
				\multicolumn{7}{l}{\textbf{\textit{Transformer-based or Advanced MIL Models}}} \\
				TransMIL(2022)~\citep{shao2021transmil}
				& 0.5519 \ensuremath{\pm} 0.0023 & 0.6934 \ensuremath{\pm} 0.0020 & 0.9489 \ensuremath{\pm} 0.0015
				& 0.9958 \ensuremath{\pm} 0.0007 & \textbf{1.0000 \ensuremath{\pm} 0.0000} & \textbf{1.0000 \ensuremath{\pm} 0.0000} \\
				IBMIL(2023)~\citep{lin2023interventional}
				& 0.5420 \ensuremath{\pm} 0.0024 & 0.7615 \ensuremath{\pm} 0.0019 & 0.8729 \ensuremath{\pm} 0.0018
				& 0.9926 \ensuremath{\pm} 0.0008 & \textbf{1.0000 \ensuremath{\pm} 0.0000} & \textbf{1.0000 \ensuremath{\pm} 0.0000} \\
				R\textsuperscript{2}T-MIL(2024)~\citep{tang2024feature}
				& \textbf{0.6857 \ensuremath{\pm} 0.0020} & 0.8230 \ensuremath{\pm} 0.0018 & 0.9106 \ensuremath{\pm} 0.0016
				& \textbf{1.0000 \ensuremath{\pm} 0.0000} & \textbf{1.0000 \ensuremath{\pm} 0.0000} & \textbf{1.0000 \ensuremath{\pm} 0.0000} \\
				DSAGL(Ours)
				& 0.6027 \ensuremath{\pm} 0.0022
				& \textcolor{red}{\textbf{0.8286 \ensuremath{\pm} 0.0016}}
				& \textcolor{red}{\textbf{0.9697 \ensuremath{\pm} 0.0011}}
				& \textcolor{red}{\textbf{1.0000 \ensuremath{\pm} 0.0000}}
				& \textcolor{red}{\textbf{1.0000 \ensuremath{\pm} 0.0000}}
				& \textcolor{red}{\textbf{1.0000 \ensuremath{\pm} 0.0000}} \\
				\bottomrule
			\end{tabular}
		\end{subtable}
	\end{table}

	\subsubsection{Results on the NCT-CRC Dataset}
	
	\rev{As shown in Table \ref{tab:crc}, DSAGL consistently outperforms all competing methods on the NCT-CRC dataset across different positive patch ratios, demonstrating strong robustness in real-world histopathological settings. At the instance level, DSAGL maintains an AUC above about 0.95 under all supervision conditions and achieves a peak performance of about 0.9898 when the positive patch ratio reaches 70\%. Notably, under the more challenging weak-supervision scenario with a 10\% positive ratio, DSAGL attains an instance-level AUC of about 0.9714, substantially exceeding R\textsuperscript{2}T-MIL (about 0.8224), IBMIL (about 0.7515), and CLAM (about 0.6508). These results indicate that DSAGL is effective at mitigating instance-level ambiguity and preserving discriminative capacity even when supervision is sparse.}
	
	\rev{As the positive patch ratio increases to 20\% and 50\%, DSAGL further improves its instance-level performance, achieving AUCs of about 0.9520 and 0.9737, respectively. In contrast, several competing methods exhibit notable performance instability under the same conditions. For example, CLAM experiences a pronounced degradation at a 50\% positive ratio, where its instance-level AUC drops to about 0.5434, while TransMIL decreases to about 0.6026. Such fluctuations highlight the sensitivity of these models to morphological heterogeneity and distributional shifts commonly observed in real histopathological data. By comparison, DSAGL shows a smooth and monotonic performance trend across all ratios, reflecting stronger robustness to tissue-level variability.}
	
	\rev{At the bag level, most methods approach performance saturation as the positive patch ratio increases. However, DSAGL achieves an AUC of about 1.0000 even at the lowest evaluated ratio of 10\%, outperforming classical and attention-based MIL methods such as MI-Net (about 0.7142) and ABMIL (about 0.7389), as well as advanced models like TransMIL (about 0.9705). For positive ratios of 20\%, 50\%, and 70\%, DSAGL consistently maintains perfect bag-level classification performance, matching or exceeding the best-performing baselines. This result suggests that DSAGL provides highly stable slide-level predictions in realistic histopathological scenarios.}
	
\begin{figure}[h]
		\centering
		\begin{subfigure}[b]{0.45\textwidth} % 调整子图宽度
			\centering
			\includegraphics[width=1.0\textwidth]{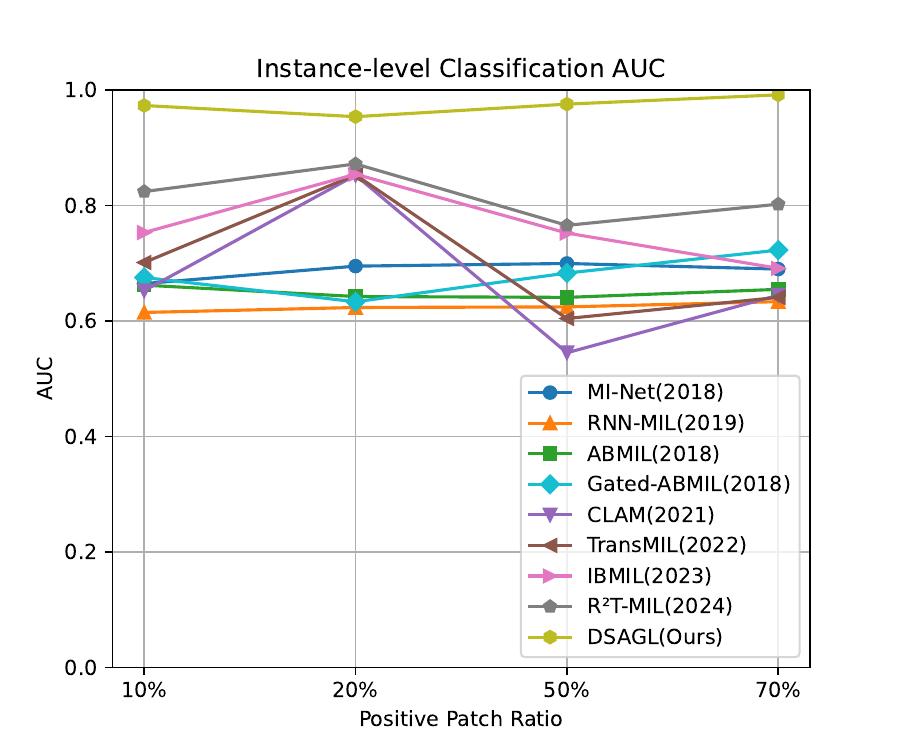} % 调整图片宽度
			\caption{Instance-level Classification AUC.}
			\label{fig:ins_auc}
		\end{subfigure}
		\hspace{15pt} % 控制两图之间的距离
		\begin{subfigure}[b]{0.45\textwidth} % 调整子图宽度
			\centering
			\includegraphics[width=1.0\textwidth]{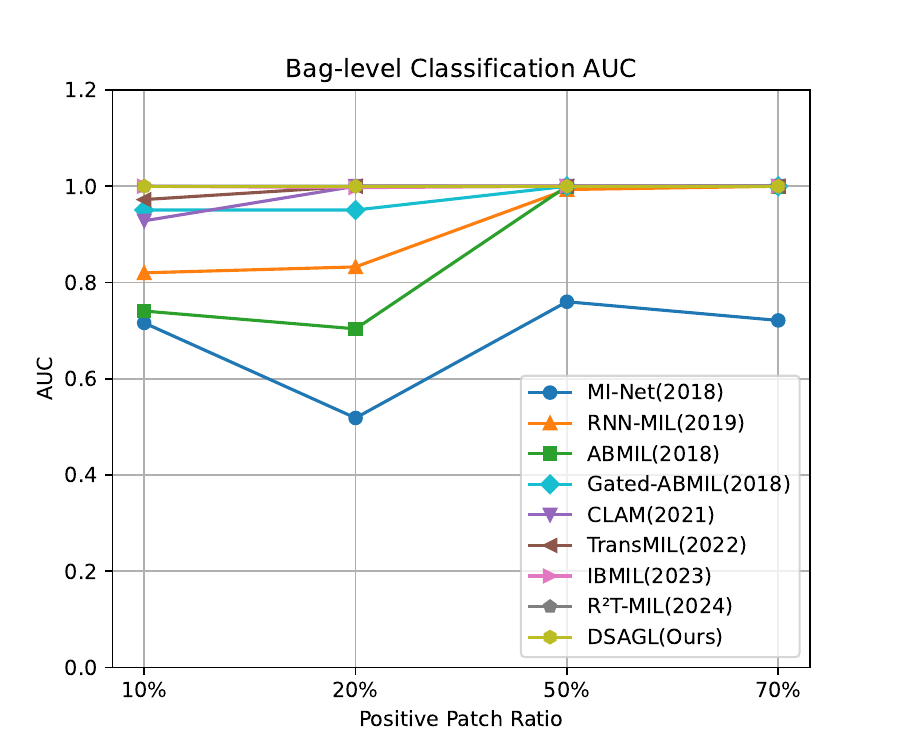} % 调整图片宽度
			\caption{Bag-level Classification AUC.}
			\label{fig:bag_auc}
		\end{subfigure}
		\caption{Comparison of AUC for Different Classification Levels on the NCT-CRC dataset.}
		\label{fig:combined_auc3}
	\end{figure}

	\rev{Figure \ref{fig:combined_auc3} further illustrates the instance-level and bag-level AUC trends across different supervision strengths. Unlike several competing approaches that exhibit noticeable degradation or variability as the supervision signal changes, DSAGL maintains consistent and reliable performance throughout the entire range. These findings confirm the effectiveness of the proposed dual-stream architecture and attention-guided pseudo-supervision strategy in handling heterogeneous tissue patterns under weak supervision, and they provide strong motivation for evaluating DSAGL on large-scale whole-slide image datasets such as TCGA-Lung.}
	
	\begin{table}[h]
		\centering
		\scriptsize
		\caption{Results on the NCT-CRC dataset.}
		\label{tab:crc}
		
		\begin{subtable}[t]{\textwidth}
			\centering
			\setlength{\tabcolsep}{6pt}
			\caption{Instance-level classification AUC.}
			\begin{tabular}{lcccc}
				\toprule
				\textbf{Positive patch ratio} & 10\% & 20\% & 50\% & 70\% \\
				\midrule
				\multicolumn{5}{l}{\textbf{\textit{Classical MIL Models}}} \\
				MI-Net(2018)~\citep{wang2018revisiting}
				& 0.6631 \ensuremath{\pm} 0.0021 & 0.6933 \ensuremath{\pm} 0.0019 & 0.6978 \ensuremath{\pm} 0.0018 & 0.6881 \ensuremath{\pm} 0.0020 \\
				RNN-MIL(2019)~\citep{campanella2019clinical}
				& 0.6131 \ensuremath{\pm} 0.0024 & 0.6217 \ensuremath{\pm} 0.0022 & 0.6226 \ensuremath{\pm} 0.0021 & 0.6317 \ensuremath{\pm} 0.0023 \\
				\midrule
				\multicolumn{5}{l}{\textbf{\textit{Attention-based MIL Models}}} \\
				ABMIL(2018)~\citep{ilse2018attention}
				& 0.6604 \ensuremath{\pm} 0.0023 & 0.6407 \ensuremath{\pm} 0.0021 & 0.6391 \ensuremath{\pm} 0.0020 & 0.6530 \ensuremath{\pm} 0.0022 \\
				Gated-ABMIL(2018)~\citep{ilse2018attention}
				& 0.6736 \ensuremath{\pm} 0.0020 & 0.6316 \ensuremath{\pm} 0.0022 & 0.6812 \ensuremath{\pm} 0.0021 & 0.7209 \ensuremath{\pm} 0.0019 \\
				CLAM(2021)~\citep{lu2021data}
				& 0.6508 \ensuremath{\pm} 0.0024 & 0.8507 \ensuremath{\pm} 0.0018 & 0.5434 \ensuremath{\pm} 0.0025 & 0.6429 \ensuremath{\pm} 0.0023 \\
				\midrule
				\multicolumn{5}{l}{\textbf{\textit{Transformer-based or Advanced MIL Models}}} \\
				TransMIL(2022)~\citep{shao2021transmil}
				& 0.6996 \ensuremath{\pm} 0.0021 & 0.8512 \ensuremath{\pm} 0.0019 & 0.6026 \ensuremath{\pm} 0.0023 & 0.6705 \ensuremath{\pm} 0.0022 \\
				IBMIL(2023)~\citep{lin2023interventional}
				& 0.7515 \ensuremath{\pm} 0.0018 & 0.8524 \ensuremath{\pm} 0.0017 & 0.7506 \ensuremath{\pm} 0.0019 & 0.8194 \ensuremath{\pm} 0.0018 \\
				R\textsuperscript{2}T-MIL(2024)~\citep{tang2024feature}
				& 0.8224 \ensuremath{\pm} 0.0016 & 0.8702 \ensuremath{\pm} 0.0015 & 0.7638 \ensuremath{\pm} 0.0018 & 0.8006 \ensuremath{\pm} 0.0017 \\
				DSAGL(Ours)
				& \textcolor{red}{\textbf{0.9714 \ensuremath{\pm} 0.0013}}
				& \textcolor{red}{\textbf{0.9520 \ensuremath{\pm} 0.0014}}
				& \textcolor{red}{\textbf{0.9737 \ensuremath{\pm} 0.0012}}
				& \textcolor{red}{\textbf{0.9898 \ensuremath{\pm} 0.0011}} \\
				\bottomrule
			\end{tabular}
		\end{subtable}
		
		\vspace{6pt} % 控制两个子表之间的竖直间距
		
		\begin{subtable}[t]{\textwidth}
			\centering
			\setlength{\tabcolsep}{6pt}
			\caption{Bag-level classification AUC.}
			\begin{tabular}{lcccc}
				\toprule
				\textbf{Positive patch ratio} & 10\% & 20\% & 50\% & 70\% \\
				\midrule
				\multicolumn{5}{l}{\textbf{\textit{Classical MIL Models}}} \\
				MI-Net(2018)~\citep{wang2018revisiting}
				& 0.7142 \ensuremath{\pm} 0.0020 & 0.5168 \ensuremath{\pm} 0.0023 & 0.7582 \ensuremath{\pm} 0.0019 & 0.7194 \ensuremath{\pm} 0.0021 \\
				RNN-MIL(2019)~\citep{campanella2019clinical}
				& 0.8183 \ensuremath{\pm} 0.0019 & 0.8307 \ensuremath{\pm} 0.0018 & 0.9915 \ensuremath{\pm} 0.0006 & \textbf{1.0000 \ensuremath{\pm} 0.0000} \\
				\midrule
				\multicolumn{5}{l}{\textbf{\textit{Attention-based MIL Models}}} \\
				ABMIL(2018)~\citep{ilse2018attention}
				& 0.7389 \ensuremath{\pm} 0.0021 & 0.7019 \ensuremath{\pm} 0.0022 & \textbf{1.0000 \ensuremath{\pm} 0.0000} & \textbf{1.0000 \ensuremath{\pm} 0.0000} \\
				Gated-ABMIL(2018)~\citep{ilse2018attention}
				& 0.9488 \ensuremath{\pm} 0.0015 & 0.9489 \ensuremath{\pm} 0.0016 & \textbf{1.0000 \ensuremath{\pm} 0.0000} & \textbf{1.0000 \ensuremath{\pm} 0.0000} \\
				CLAM(2021)~\citep{lu2021data}
				& 0.9811 \ensuremath{\pm} 0.0012 & \textbf{1.0000 \ensuremath{\pm} 0.0000} & \textbf{1.0000 \ensuremath{\pm} 0.0000} & \textbf{1.0000 \ensuremath{\pm} 0.0000} \\
				\midrule
				\multicolumn{5}{l}{\textbf{\textit{Transformer-based or Advanced MIL Models}}} \\
				TransMIL(2022)~\citep{shao2021transmil}
				& 0.9705 \ensuremath{\pm} 0.0013 & \textbf{1.0000 \ensuremath{\pm} 0.0000} & \textbf{1.0000 \ensuremath{\pm} 0.0000} & \textbf{1.0000 \ensuremath{\pm} 0.0000} \\
				IBMIL(2023)~\citep{lin2023interventional}
				& \textbf{1.0000 \ensuremath{\pm} 0.0000} & 0.9961 \ensuremath{\pm} 0.0007 & \textbf{1.0000 \ensuremath{\pm} 0.0000} & \textbf{1.0000 \ensuremath{\pm} 0.0000} \\
				R\textsuperscript{2}T-MIL~\citep{tang2024feature}
				& \textbf{1.0000 \ensuremath{\pm} 0.0000} & \textbf{1.0000 \ensuremath{\pm} 0.0000} & 0.9980 \ensuremath{\pm} 0.0008 & \textbf{1.0000 \ensuremath{\pm} 0.0000} \\
				DSAGL(Ours)
				& \textcolor{red}{\textbf{1.0000 \ensuremath{\pm} 0.0000}}
				& \textcolor{red}{\textbf{1.0000 \ensuremath{\pm} 0.0000}}
				& \textcolor{red}{\textbf{1.0000 \ensuremath{\pm} 0.0000}}
				& \textcolor{red}{\textbf{1.0000 \ensuremath{\pm} 0.0000}} \\
				\bottomrule
			\end{tabular}
		\end{subtable}
	\end{table}

	\subsubsection{Results on the TCGA-Lung dataset}
	
	\rev{On the real-world TCGA-Lung dataset, we further evaluate the proposed method on bag-level classification for gigapixel whole-slide images (WSIs), with the results summarized in Table \ref{tab:tcga}. From the overall comparison, DSAGL achieves the highest bag-level AUC of about 0.9706 on the TCGA-Lung dataset, outperforming all competing methods. Specifically, among classical MIL approaches, MI-Net and RNN-MIL obtain bag-level AUCs of around 0.9291 and 0.9284, respectively, demonstrating a certain level of discriminative capability under complex pathological backgrounds, but remaining limited by their coarse-grained feature aggregation mechanisms. Attention-based methods show improved performance on this dataset, with ABMIL and Gated-ABMIL achieving AUCs of approximately 0.9466 and 0.9564, respectively, while CLAM further improves performance to about 0.9589, highlighting the effectiveness of attention-based instance selection in WSI scenarios.}
	
	\rev{By comparison, Transformer-based and advanced MIL methods exhibit more robust performance overall. TransMIL and IBMIL achieve bag-level AUCs of around 0.9653 and 0.9671, respectively, and the recent R\textsuperscript{2}T-MIL further improves performance to approximately 0.9685. Building upon these strong baselines, DSAGL consistently attains a slightly higher bag-level AUC of about 0.9706 by introducing a dual-stream collaborative learning framework and attention-guided pseudo-supervision, without relying on additional fine-grained annotations. Although the numerical improvement is relatively modest, on a highly heterogeneous and large-scale real WSI dataset such as TCGA-Lung, this gain reflects a consistent advantage in overall discriminative stability and feature modeling capability.}
	
	\rev{These results indicate that DSAGL can effectively integrate global semantic information with local discriminative cues in real clinical pathology data, enabling reliable slide-level classification under weak supervision. This performance trend is consistent with the instance-level stability observed on the NCT-CRC dataset, further validating the generalization capability of the proposed method across pathological scenarios with different scales and levels of complexity, and supporting its potential applicability in practical computational pathology workflows.}
	
	\begin{table}[h]
		\centering
		\scriptsize
		\caption{Results on the TCGA-Lung dataset.}
		\label{tab:tcga}
		\begin{tabular}{lc}
			\toprule
			\textbf{Method} & \textbf{Bag-level AUC} \\
			\midrule
			\multicolumn{2}{l}{\textbf{\textit{Classical MIL Models}}} \\
			MI-Net(2018)~\citep{wang2018revisiting} & 0.9291 \ensuremath{\pm} 0.0032 \\
			RNN-MIL(2019)~\citep{campanella2019clinical} & 0.9284 \ensuremath{\pm} 0.0035 \\
			\midrule
			\multicolumn{2}{l}{\textbf{\textit{Attention-based MIL Models}}} \\
			ABMIL(2018)~\citep{ilse2018attention} & 0.9466 \ensuremath{\pm} 0.0028 \\
			Gated-ABMIL(2018)~\citep{ilse2018attention} & 0.9564 \ensuremath{\pm} 0.0026 \\
			CLAM(2021)~\citep{lu2021data} & 0.9589 \ensuremath{\pm} 0.0024 \\
			\midrule
			\multicolumn{2}{l}{\textbf{\textit{Transformer-based or Advanced MIL Models}}} \\
			TransMIL(2021)~\citep{shao2021transmil} & 0.9653 \ensuremath{\pm} 0.0021 \\
			IBMIL(2023)~\citep{lin2023interventional} & 0.9671 \ensuremath{\pm} 0.0020 \\
			R\textsuperscript{2}T-MIL(2024)~\citep{tang2024feature} & 0.9685 \ensuremath{\pm} 0.0019 \\
			DSAGL(Ours) & \textcolor{red}{\textbf{0.9706 \ensuremath{\pm} 0.0018}} \\
			\bottomrule
		\end{tabular}
	\end{table}
	\subsection{Ablation Study}
	\rev{\subsubsection{Backbone Ablation and Efficiency Analysis}}
	\rev{We perform a backbone ablation and efficiency analysis on TCGA-Lung to study the effect of the encoder design and to quantify its computational cost. All models are evaluated under the same MIL pipeline and experimental settings. Specifically, each whole-slide image is represented by a fixed-size bag of $K=256$ patch features during inference, and both latency and peak GPU memory are measured on a per-slide basis without backpropagation. Under this controlled setup, we compare four backbones: a CNN-only baseline without Mamba, DeiT-Ti~\citep{touvron2021training} as a lightweight Transformer alternative, Vim-Ti~\citep{vim} as a Vision Mamba baseline, and the proposed VSSMamba.}
	
	\rev{As shown in Table~\ref{tab:backbone_ablation_main}, VSSMamba achieves the best Bag AUC of about 0.9706, outperforming CNN-only (about 0.9151), DeiT-Ti (about 0.9673), and Vim-Ti (about 0.9680). At the same time, VSSMamba remains lightweight, with about 6.94M parameters, which is comparable to Vim-Ti (about 7.00M) and smaller than the CNN-only baseline (about 8.50M). In terms of efficiency, VSSMamba requires about 1.50 GFLOPs per slide, about 0.40 GB peak memory, and about 0.010 s/slide latency, compared with about 1.20 GFLOPs / 0.38 GB / 0.012 s for CNN-only, about 6.50 GFLOPs / 0.65 GB / 0.040 s for DeiT-Ti, and about 9.10 GFLOPs / 0.75 GB / 0.050 s for Vim-Ti. These results indicate that the proposed backbone achieves a more favorable accuracy–efficiency trade-off than both CNN-only and alternative sequence-based backbones.}
	
	\rev{We further evaluate scalability at a larger bag size of $K=1024$ in Table~\ref{tab:backbone_scalability}. As the number of instances per slide increases, Transformer-based backbones exhibit a rapid growth in both memory consumption and inference time. In contrast, VSSMamba scales more favorably, requiring about 0.95 GB peak memory and about 0.040 s/slide latency, compared with about 2.60 GB / 0.160 s for DeiT-Ti and about 3.10 GB / 0.190 s for Vim-Ti. This demonstrates that the proposed backbone maintains practical efficiency when processing slides with larger numbers of instances, which is critical for real-world WSI analysis. Note that peak GPU memory does not scale strictly linearly with $K$, since inference does not retain intermediate activations and a large portion of the memory footprint comes from constant model buffers.}
	
	\begin{table}[t]
		\centering
		\scriptsize
		\setlength{\tabcolsep}{5pt}
		\caption{\rev{Backbone ablation and efficiency analysis on TCGA-Lung.}}
		\label{tab:backbone_ablation_main}
		\begin{tabular}{l|c|cccc}
			\toprule
			Backbone
			& Bag AUC
			& Params (M)
			& GFLOPs/slide
			& Peak Mem (GB)
			& Latency (s/slide) \\
			\midrule
			CNN-only
			& 0.9151 \ensuremath{\pm} 0.0197
			& 8.50
			& \textbf{1.20}
			& \textbf{0.38}
			& 0.012 \\
			
			DeiT-Ti~\citep{touvron2021training}
			& 0.9673 \ensuremath{\pm} 0.0206
			& 5.00
			& 6.50
			& 0.65
			& 0.040 \\
			
			Vim-Ti~\citep{vim}
			& 0.9680 \ensuremath{\pm} 0.0091
			& 7.00
			& 9.10
			& 0.75
			& 0.050 \\
			
			VSSMamba (Ours)
			& \textcolor{red}{\textbf{0.9706 \ensuremath{\pm} 0.0018}}
			& \textcolor{red}{\textbf{6.94}}
			& 1.50
			& 0.40
			& \textcolor{red}{\textbf{0.010}} \\
			\bottomrule
		\end{tabular}
	\end{table}
	
	\begin{table}[t]
		\centering
		\scriptsize
		\setlength{\tabcolsep}{6pt}
		\caption{\rev{Scalability comparison on TCGA-Lung at bag size $K=1024$.}}
		\label{tab:backbone_scalability}
		\begin{tabular}{l|cc}
			\toprule
			Backbone
			& Peak Mem (GB)
			& Latency (s/slide)\\
			\midrule
			CNN-only               & 0.97 & 0.044 \\
			DeiT-Ti~\citep{touvron2021training}  & 2.60 & 0.160 \\
			Vim-Ti~\citep{vim}  & 3.10 & 0.190 \\
			VSSMamba (Ours)  & \textcolor{red}{\textbf{0.95}} & \textcolor{red}{\textbf{0.040}} \\
			\bottomrule
		\end{tabular}
	\end{table}
	
	\rev{\subsubsection{Component-wise Architecture Design}}
	
	To systematically evaluate the contribution of each core component within the DSAGL to classification performance at different granularities, we conducted an ablation study on the NCT-CRC dataset with a positive instance ratio of 10\%. The results are summarized in Table~\ref{tab:ablation}.
	
	\begin{table}[h]
		\centering
		\scriptsize
		\setlength{\tabcolsep}{16pt}
		\caption{Ablation study on the NCT-CRC dataset.}
		\label{tab:ablation}
		\begin{tabular}{cccccc}
			\toprule
			\textbf{Dual-stream} & \textbf{FASA} & $\mathcal{L}_{Hybrid}$ & \textbf{VSSMamba} & \textbf{Instance AUC} & \textbf{Bag AUC} \\
			\midrule
			& & & & 0.7392 \ensuremath{\pm} 0.0031 & 0.9736 \ensuremath{\pm} 0.0012 \\
			\checkmark & & & & 0.8003 \ensuremath{\pm} 0.0026 & \textbf{1.0000 \ensuremath{\pm} 0.0000} \\
			\checkmark & \checkmark & & & 0.9196 \ensuremath{\pm} 0.0019 & \textbf{1.0000 \ensuremath{\pm} 0.0000} \\
			\checkmark & \checkmark & \checkmark & & 0.9330 \ensuremath{\pm} 0.0017 & \textbf{1.0000 \ensuremath{\pm} 0.0000} \\
			\checkmark & \checkmark & \checkmark & \checkmark
			& \textcolor{red}{\textbf{0.9714 \ensuremath{\pm} 0.0013}}
			& \textcolor{red}{\textbf{1.0000 \ensuremath{\pm} 0.0000}} \\
			\bottomrule
		\end{tabular}
	\end{table}
	
	Starting from a baseline model without any additional modules, we observe an instance-level AUC of about 0.7392 and a bag-level AUC of about 0.9736. Introducing the dual-stream architecture improves the instance-level AUC to about 0.8003, while the bag-level AUC reaches about 1.0000—indicating that the teacher–student collaborative optimization mechanism effectively enhances bag-level prediction performance. Incorporating the FASA module further boosts the instance-level AUC to about 0.9196, demonstrating the effectiveness of multi-scale saliency modeling in identifying diagnostically relevant lesion regions.
	
	Building on this, the addition of the hybrid loss function $\mathcal{L}_{Hybrid}$ leads to a further increase in instance-level AUC to about 0.9330, highlighting the robustness advantages of joint supervision through both hard and soft pseudo labels under weak supervision. Finally, integrating the customized feature encoder VSSMamba into the overall architecture yields the best performance, achieving an instance-level AUC of about 0.9714 while maintaining a bag-level AUC of about 1.0000. This result validates the effectiveness of VSSMamba in modeling long-range dependencies and integrating multi-scale features for enhanced instance-level discrimination.
	
	\rev{While the above ablation study verifies the effectiveness of incorporating VSSMamba into the DSAGL framework, the optimal depth of Mamba layers may vary across datasets with different instance resolutions and semantic complexities. To clarify the hyperparameter selection process and improve reproducibility, we further investigate the impact of the number of Mamba layers on model performance. Table~\ref{tab:mamba_depth} summarizes the performance of VSSMamba under different Mamba layer configurations on CIFAR-10, NCT-CRC, and TCGA-Lung.}
	
	\begin{table}[h]
		\centering
		\scriptsize
		\setlength{\tabcolsep}{12pt}
		\caption{\rev{Effect of Mamba layer depth in VSSMamba across different datasets.}}
		\label{tab:mamba_depth}
		\begin{tabular}{ccccc}
			\toprule
			\textbf{Dataset} & \textbf{Mamba Layers} & \textbf{Positive Patch Ratio (\%)} & \textbf{Instance-level AUC} & \textbf{Bag-level AUC} \\
			\midrule
			CIFAR-10   & 0 & 20 & \textcolor{red}{\textbf{0.9268 \ensuremath{\pm} 0.0013}} & \textcolor{red}{\textbf{1.0000 \ensuremath{\pm} 0.0000}} \\
			CIFAR-10   & 1 & 20 & 0.8916 \ensuremath{\pm} 0.0011 & \textbf{1.0000 \ensuremath{\pm} 0.0000} \\
			\midrule
			NCT-CRC    & 1 & 20 & 0.9291 \ensuremath{\pm} 0.0021 & 0.9962 \ensuremath{\pm} 0.0010 \\
			NCT-CRC    & 2 & 20 & \textcolor{red}{\textbf{0.9520 \ensuremath{\pm} 0.0014}} & \textcolor{red}{\textbf{1.0000 \ensuremath{\pm} 0.0000}} \\
			NCT-CRC    & 3 & 20 & 0.9400 \ensuremath{\pm} 0.0012 & \textbf{1.0000 \ensuremath{\pm} 0.0000} \\
			\midrule
			TCGA-Lung  & 2 & -- & -- & 0.9601 \ensuremath{\pm} 0.0019 \\
			TCGA-Lung  & 3 & -- & -- & \textcolor{red}{\textbf{0.9706 \ensuremath{\pm} 0.0018}} \\
			TCGA-Lung  & 4 & -- & -- & 0.9691 \ensuremath{\pm} 0.0011 \\
			\bottomrule
		\end{tabular}
	\end{table}
	
	\rev{For CIFAR-10, which consists of low-resolution instances, introducing Mamba layers does not yield performance gains, and the best results are consistently achieved with zero Mamba layers. In contrast, for the NCT-CRC dataset with higher instance resolution and richer spatial structures, using two Mamba layers achieves the best trade-off between performance and model complexity, reaching an instance-level AUC of about 0.9520 and a bag-level AUC of about 1.0000. For the gigapixel-scale TCGA-Lung dataset, three Mamba layers consistently provide the highest bag-level AUC of about 0.9706, while deeper configurations do not lead to further improvements.}
	
	\rev{Based on these observations, we adopt dataset-specific Mamba layer configurations in our final model design, setting the number of Mamba layers to 0, 2, and 3 for CIFAR-10, NCT-CRC, and TCGA-Lung, respectively.}
	
	\rev{\subsubsection{Training Strategy Analysis}}
	\rev{To investigate the impact of different optimization strategies on weakly supervised learning, we conduct a controlled comparison among three representative training paradigms: sequential training, joint training, and the proposed alternating training strategy. All experiments are performed on the NCT-CRC dataset under a fixed positive patch ratio of 20\%, while keeping the model architecture, loss functions, and hyperparameters identical across different settings.}
	
	\rev{As summarized in Table~\ref{tab:training_strategy}, sequential training, which optimizes the teacher and student streams in separate stages, yields a relatively low instance-level AUC of about 0.8287, despite achieving a perfect bag-level AUC. This result indicates that decoupling the optimization process may limit the effectiveness of knowledge transfer from the bag-level teacher to the instance-level student, leading to suboptimal patch-level discrimination.}
	
	\rev{Joint training improves the instance-level AUC to about 0.9245 by simultaneously updating both streams within each iteration, enabling direct supervision and tighter coupling between the teacher and student. However, such a strategy requires concurrent backpropagation through both streams, which substantially increases GPU memory consumption and may introduce optimization instability under large instance batches, as observed in our experiments.}
	
	\rev{In contrast, the proposed alternating training strategy achieves the best instance-level performance, reaching an AUC of about 0.9520, while consistently maintaining a perfect bag-level AUC. By alternately optimizing the teacher and student streams, this strategy effectively decouples their gradient updates, thereby reducing gradient interference and mitigating the propagation of noisy pseudo labels. At the same time, it preserves sufficient inter-stream interaction to facilitate stable and efficient knowledge transfer.}
	
	\rev{Overall, these results demonstrate that alternating optimization offers a favorable balance between performance, stability, and practicality, validating it as a principled training strategy rather than a mere implementation choice.}
	
	\begin{table}[h]
		\centering
		\scriptsize
		\setlength{\tabcolsep}{10pt}
		\caption{\rev{Comparison of different training strategies on the NCT-CRC dataset under a 20\% positive patch ratio setting.}}
		\label{tab:training_strategy}
		\begin{tabular}{lcc}
			\toprule
			\textbf{Training Strategy} & \textbf{Instance AUC} & \textbf{Bag AUC} \\
			\midrule
			Sequential Training & 0.8287 \ensuremath{\pm} 0.0010 & \textbf{1.0000 \ensuremath{\pm} 0.0000} \\
			Joint Training      & 0.9245 \ensuremath{\pm} 0.0112 & \textbf{1.0000 \ensuremath{\pm} 0.0000} \\
			Alternating (Ours)  & \textcolor{red}{\textbf{0.9520 \ensuremath{\pm} 0.0014}} & \textcolor{red}{\textbf{1.0000 \ensuremath{\pm} 0.0000}} \\
			\bottomrule
		\end{tabular}
	\end{table}
		
	\subsubsection{Loss Function Design}
	
	To further investigate the contribution of each component in our loss function design, we conduct a series of ablation experiments on the CIFAR-10 dataset with a positive patch ratio of 20\%, as summarized in Table~\ref{tab:loss_ablation}. These experiments focus on three key factors: whether the binary cross-entropy (BCE) loss is class-balanced, whether knowledge distillation via KL divergence is applied, and what type of pseudo-label is utilized during training.
	
	Under the full configuration (which includes weighted BCE, KL divergence, and soft pseudo-labels), our method achieves the best performance, with an instance-level AUC of about 0.9268 and a bag-level AUC of about 1.0000. Removing the KL divergence term leads to a noticeable drop in instance AUC to about 0.9099, indicating that mimicking the teacher's soft outputs significantly enhances the student's representational learning.
	
	Further eliminating the class-balancing mechanism in BCE results in a more pronounced performance degradation: the instance AUC decreases to about 0.8813 and the bag-level AUC falls to about 0.9796. This suggests that class imbalance adversely affects optimization, and that the reweighted BCE term contributes to better training stability.
	
	When the soft pseudo-labels are binarized into hard labels, the instance AUC declines to about 0.8906, even though KL divergence is retained. This underscores the importance of preserving uncertainty information in the pseudo-labels to guide the student more effectively. Moreover, when pseudo-labels are randomly assigned, performance drops drastically, with the instance AUC and bag-level AUC plunging to about 0.6108 and about 0.6731, respectively, highlighting the critical role of pseudo-label quality in weakly supervised learning.
	
	As a reference upper bound, we also evaluate a student model trained with ground-truth instance labels, which yields an instance-level AUC of about 0.9203 and a bag-level AUC of about 1.0000. While slightly lower than the full hybrid loss configuration, this result demonstrates that our pseudo-labeling strategy is already approaching the performance of fully supervised training.
	
	Overall, these findings confirm that the proposed hybrid loss, which integrates confidence-aware supervision, class-balanced reweighting, and soft-label distillation, is critical for achieving robust and generalizable learning under weak supervision.	
	\begin{table}[h]
		\centering
		\scriptsize
		\renewcommand{\arraystretch}{1.2}
		\setlength{\tabcolsep}{8pt}
		\caption{Ablation study on the effects of pseudo-label design, KL divergence, and weighted BCE on the CIFAR-10 dataset.}
		\label{tab:loss_ablation}
		\begin{tabular}{cccccc}
			\toprule
			\textbf{Weighted BCE} & \textbf{KL Divergence} & \textbf{Pseudo-label Type} & \textbf{Instance AUC} & \textbf{Bag AUC} \\
			\midrule
			\checkmark & \checkmark & Soft
			& \textcolor{red}{\textbf{0.9268 \ensuremath{\pm} 0.0013}}
			& \textcolor{red}{\textbf{1.0000 \ensuremath{\pm} 0.0000}} \\
			\checkmark &            & Soft
			& 0.9099 \ensuremath{\pm} 0.0017
			& \textbf{1.0000 \ensuremath{\pm} 0.0000} \\
			&            & Soft
			& 0.8813 \ensuremath{\pm} 0.0021
			& 0.9796 \ensuremath{\pm} 0.0013 \\
			\checkmark & \checkmark & Hard
			& 0.8906 \ensuremath{\pm} 0.0018
			& \textbf{1.0000 \ensuremath{\pm} 0.0000} \\
			\checkmark &            & Random
			& 0.6108 \ensuremath{\pm} 0.0025
			& 0.6731 \ensuremath{\pm} 0.0022 \\
			\checkmark &            & GT
			& 0.9203 \ensuremath{\pm} 0.0015
			& \textbf{1.0000 \ensuremath{\pm} 0.0000} \\
			\bottomrule
		\end{tabular}
	\end{table}
	
	\rev{\subsubsection{Hyperparameter Sensitivity Analysis}}
	\rev{To assess the robustness of the proposed mixed loss, we conduct a hyperparameter sensitivity analysis on the CE--KL balance parameter $\alpha$, the temperature $T$, and the negative weighting factor $w_n$. In each experiment, only one hyperparameter is varied while all other hyperparameters are fixed to their best-performing values. The results on the NCT-CRC dataset with a 20\% positive patch ratio are summarized in Table~\ref{tab:hyperparameter_sensitivity}.}
	
	\rev{As shown in Table~\ref{tab:hyperparameter_sensitivity}(a) and (b), the proposed method exhibits stable performance across a broad range of $\alpha$ and $T$ values. Increasing $\alpha$ from 0.60 to 0.90 consistently improves instance-level AUC, while further increasing it to 1.00 yields no additional gains, indicating that an appropriate balance between supervised learning and distillation is beneficial. Similarly, varying the temperature $T$ leads to only minor performance fluctuations, with the default setting $T=1.0$ achieving the best instance-level performance. In all cases, the bag-level AUC remains saturated, suggesting that the overall slide-level prediction is robust to these hyperparameters.}
	
	\rev{Table~\ref{tab:hyperparameter_sensitivity}(c) reports the sensitivity to the negative weighting factor $w_n$. The instance-level performance peaks at $w_n = 0.10$ and degrades smoothly when deviating from this value, reflecting the importance of down-weighting noisy negative samples in weakly supervised learning. Nevertheless, the absence of abrupt performance drops indicates that the proposed framework is not brittle with respect to the choice of $w_n$. Overall, these results demonstrate that the selected hyperparameter settings provide a favorable trade-off between performance and robustness.}
	\begin{table}[t]
		\centering
		\scriptsize
		\caption{\rev{Hyperparameter sensitivity analysis of the mixed loss function on the NCT-CRC dataset (20\% positive patch ratio).}}
		\label{tab:hyperparameter_sensitivity}
		
		% ---------- (a) alpha ----------
		\begin{subtable}[t]{0.48\textwidth}
			\centering
			\setlength{\tabcolsep}{6pt}
			\caption{Sensitivity to $\alpha$.}
			\begin{tabular}{lcc}
				\toprule
				$\boldsymbol{\alpha}$ & Inst. AUC & Bag AUC \\
				\midrule
				0.60 & 0.9318 \ensuremath{\pm} 0.0024 & 0.9987 \ensuremath{\pm} 0.0006 \\
				0.70 & 0.9405 \ensuremath{\pm} 0.0019 & 0.9991 \ensuremath{\pm} 0.0008 \\
				0.80 & 0.9472 \ensuremath{\pm} 0.0016 & \textbf{1.0000 \ensuremath{\pm} 0.0000} \\
				0.90 & 0.9501 \ensuremath{\pm} 0.0013 & \textbf{1.0000 \ensuremath{\pm} 0.0000} \\
				1.00 & 0.9489 \ensuremath{\pm} 0.0015 & \textbf{1.0000 \ensuremath{\pm} 0.0000} \\
				\textbf{0.90$\rightarrow$1.00} & \textcolor{red}{\textbf{0.9520 \ensuremath{\pm} 0.0014}} & \textcolor{red}{\textbf{1.0000 \ensuremath{\pm} 0.0000}} \\
				\bottomrule
			\end{tabular}
		\end{subtable}
		
		\vspace{6pt}
		
		% ---------- (b) T ----------
		\begin{subtable}[t]{0.48\textwidth}
			\centering
			\setlength{\tabcolsep}{6pt}
			\caption{Sensitivity to temperature $T$.}
			\begin{tabular}{lcc}
				\toprule
				$\boldsymbol{T}$ & Inst. AUC & Bag AUC \\
				\midrule
				0.50 & 0.9440 \ensuremath{\pm} 0.0018 & 0.9963 \ensuremath{\pm} 0.0024 \\
				\textbf{1.00} & \textcolor{red}{\textbf{0.9520 \ensuremath{\pm} 0.0014}} & \textcolor{red}{\textbf{1.0000 \ensuremath{\pm} 0.0000}} \\
				2.00 & 0.9493 \ensuremath{\pm} 0.0015 & 0.9967 \ensuremath{\pm} 0.0022 \\
				4.00 & 0.9416 \ensuremath{\pm} 0.0021 & 0.9860 \ensuremath{\pm} 0.0030 \\
				\bottomrule
			\end{tabular}
		\end{subtable}
		\hfill
		% ---------- (c) w_n ----------
		\begin{subtable}[t]{0.48\textwidth}
			\centering
			\setlength{\tabcolsep}{6pt}
			\caption{Sensitivity to negative weighting $w_n$.}
			\begin{tabular}{lcc}
				\toprule
				$\boldsymbol{w_n}$ & Inst. AUC & Bag AUC \\
				\midrule
				0.05 & 0.9468 \ensuremath{\pm} 0.0017 & \textbf{1.0000 \ensuremath{\pm} 0.0000} \\
				\textbf{0.10} & \textcolor{red}{\textbf{0.9520 \ensuremath{\pm} 0.0014}} & \textcolor{red}{\textbf{1.0000 \ensuremath{\pm} 0.0000}} \\
				0.20 & 0.9485 \ensuremath{\pm} 0.0016 & \textbf{1.0000 \ensuremath{\pm} 0.0000} \\
				0.50 & 0.9399 \ensuremath{\pm} 0.0020 & 0.9991 \ensuremath{\pm} 0.0004 \\
				\bottomrule
			\end{tabular}
		\end{subtable}
		
	\end{table}

	\subsection{Visualization and Analysis}
	\rev{\subsubsection{Grad-CAM Visualization}}
	
	To further interpret the spatial reasoning behavior of DSAGL and compare its spatial reasoning behavior and saliency patterns with representative MIL models, we visualize class-discriminative saliency responses from the student \rev{stream}. Specifically, Grad-CAM is applied to the final convolutional layer of the student encoder, where gradients are backpropagated from the classification logits to generate saliency heatmaps. These visualizations highlight the regions the model attends to during inference and provide insight into its learned instance-level representations.
	
\begin{figure}[h]
		\centering
		\includegraphics[width=0.85\linewidth]{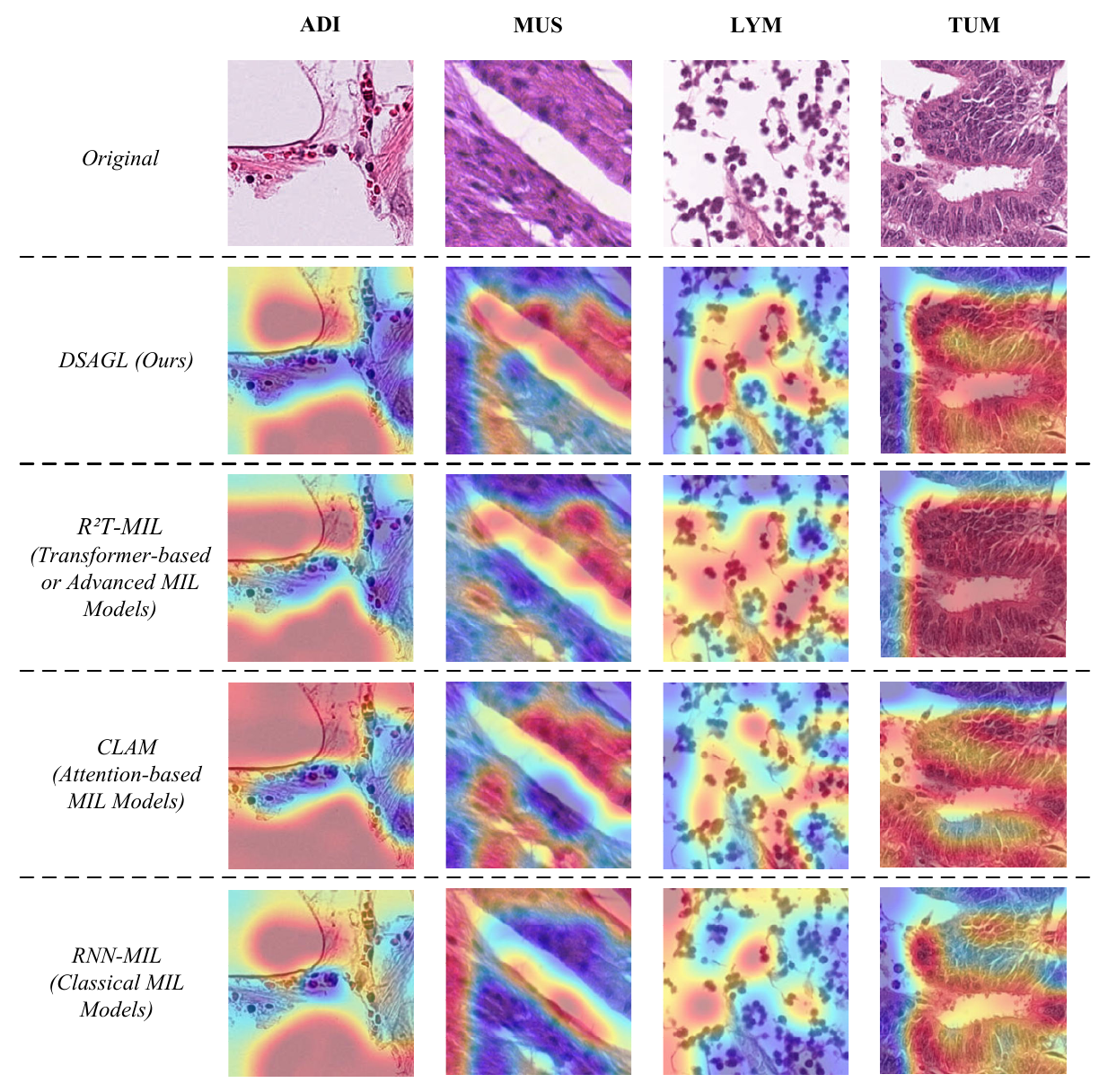}
		\caption{
			Qualitative comparison of Grad-CAM saliency maps on four tissue classes (ADI, MUS, LYM, TUM) from the NCT-CRC dataset.
			Each row corresponds to a representative MIL model: DSAGL (Ours), R$^2$T-MIL (Transformer-based), CLAM (Attention-based), and RNN-MIL (Classical pooling-based).
			DSAGL produces more localized and diagnosis-relevant Grad-CAM responses across all tissue types.
		}
		\label{fig:cam}
	\end{figure}

	Fig.~\ref{fig:cam} presents a qualitative comparison of Grad-CAM saliency heatmaps across four representative tissue classes---ADI, MUS, LYM, and TUM from the NCT-CRC dataset. The top row displays the original histopathology images, followed by visualizations from four MIL models representing distinct architectural paradigms: DSAGL (ours, representing Transformer-based or advanced MIL), R$^2$T-MIL (recent Transformer-based MIL), CLAM (attention-based MIL), and RNN-MIL (classical MIL with pooling). This selection enables a cross-generational comparison of spatial reasoning capabilities among MIL designs.
	
	Across all tissue types, DSAGL generates Grad-CAM saliency heatmaps that are consistently more focused, compact, and semantically aligned with diagnostic regions. In the ADI class (adipose tissue), DSAGL effectively suppresses homogeneous background and highlights interstitial boundaries, whereas other methods yield dispersed or poorly localized activation. For MUS (muscle), DSAGL sharply delineates fiber margins, while R$^2$T-MIL and CLAM produce broader, less discriminative responses. In the LYM class (lymphocyte) , which is challenging due to its sparse distribution, DSAGL accurately localizes small lymphocyte clusters, outperforming CLAM's overextended focus and RNN-MIL's weak saliency. For TUM (tumor), DSAGL concentrates on epithelial abnormalities and distorted glandular structures with superior boundary precision.
	
	\rev{Although DSAGL generally produces more concentrated and diagnostically relevant attention responses, a limited number of challenging cases still exhibit mild attention drift. As illustrated in Fig.~\ref{fig:cam}, for certain MUS (muscle tissue) samples, the Grad-CAM saliency maps accurately highlight the dominant muscle fiber structures but also slightly extend into adjacent connective tissue regions. This behavior is likely attributable to local pseudo-label noise introduced during the early stages of training, before the pseudo supervision fully converges, leading to minor response diffusion near tissue boundaries. A similar phenomenon can be observed in the LYM (lymphocyte) class, where the small size and highly sparse spatial distribution of lymphocyte clusters may cause weak activation responses in some non-lesion background regions. In contrast, R$^2$T-MIL and CLAM tend to produce more diffuse and spatially overextended saliency patterns in these categories, making it difficult to distinguish lesion regions from background, while RNN-MIL exhibits generally weak and less discriminative responses. Overall, these failure cases are primarily associated with scenarios involving ambiguous tissue boundaries or extremely sparse lesion distributions, reflecting the inherent challenges of attention learning under weak supervision without pixel- or region-level annotations. Nevertheless, compared with competing methods, DSAGL maintains a relatively focused response on the principal discriminative structures even in such difficult cases, indicating that the proposed alternating teacher–student optimization and attention-guided pseudo-supervision strategy effectively mitigates attention drift in practice.}
	
	Overall, the alternating teacher-student optimization and hybrid supervision strategy in DSAGL effectively reduce attention drift. Through iterative pseudo-label refinement and instance-level consistency enforcement, the model achieves more stable and diagnostically meaningful attention localization. These interpretable saliency patterns further demonstrate DSAGL's applicability and reliability in real-world weakly supervised histopathology tasks.

	\rev{\subsubsection{Quantitative Faithfulness Analysis of Saliency Maps}}

	\rev{To quantitatively complement the Grad-CAM visualizations in Fig.~\ref{fig:cam}, we evaluate the faithfulness of saliency maps on the NCT-CRC test set. All experiments are conducted at inference time using the trained student classifier, without any additional fine-tuning. We evaluate all methods on the same set of 10 test patches and report the average results. For each patch, a normalized Grad-CAM saliency map is generated and image regions are ranked according to saliency intensity. The top 30\% ranked regions are regarded as highly salient. Based on this ranking, deletion and insertion tests are performed by progressively perturbing or restoring the most salient regions. The prediction confidence at each step is defined as the softmax probability of the ground-truth class.}

	\rev{The quantitative results are summarized in Table~\ref{tab:faithfulness_analysis}. Overall, classical MIL (RNN-MIL) yields the weakest faithfulness performance, with a relatively high Deletion AUC (0.6842), low Insertion AUC (0.4127), and low CDR (0.2764), indicating limited reliance on the highlighted regions. Attention-based CLAM and Transformer-based R$^2$T-MIL show consistent improvements, suggesting that stronger representation modeling leads to more informative saliency responses. In contrast, DSAGL achieves the best performance across all three metrics, with the lowest Deletion AUC (0.5739), the highest Insertion AUC (0.5186), and the highest CDR (0.3914). Compared with R$^2$T-MIL, DSAGL further reduces Deletion AUC by 0.0476 and increases Insertion AUC and CDR by 0.0474 and 0.0536, respectively. These improvements indicate that the salient regions identified by DSAGL are more causally relevant to the final prediction.}

	\rev{In summary, the quantitative faithfulness analysis corroborates the qualitative observations in Fig.~\ref{fig:cam}. DSAGL consistently produces saliency maps whose highlighted regions are more decision-critical under both deletion and insertion perturbations. This demonstrates that the proposed alternating teacher--student optimization enables DSAGL to learn more faithful and decision-consistent saliency patterns, thereby enhancing interpretability in weakly supervised histopathology classification.}
	
	\begin{table}[h]
		
		\centering
		
		\scriptsize
		
		\setlength{\tabcolsep}{10pt}
		
		\caption{\rev{Quantitative faithfulness analysis of Grad-CAM saliency maps on the NCT-CRC dataset. Lower Deletion AUC and higher Insertion AUC or Confidence Drop Ratio indicate more faithful and decision-critical saliency regions.}}
		
		\label{tab:faithfulness_analysis}
		
		\begin{tabular}{lccc}
			
			\toprule
			
			\textbf{Method}
			& \textbf{Deletion AUC}
			& \textbf{Insertion AUC}
			& \textbf{Confidence Drop Ratio} \\
			
			\midrule
			
			RNN-MIL
			& 0.6842 \ensuremath{\pm} 0.0123
			& 0.4127 \ensuremath{\pm} 0.0108
			& 0.2764 \ensuremath{\pm} 0.0145 \\
			
			CLAM
			& 0.6428 \ensuremath{\pm} 0.0109
			& 0.4536 \ensuremath{\pm} 0.0121
			& 0.3189 \ensuremath{\pm} 0.0127 \\
			
			R$^2$T-MIL
			& 0.6215 \ensuremath{\pm} 0.0096
			& 0.4712 \ensuremath{\pm} 0.0114
			& 0.3378 \ensuremath{\pm} 0.0132 \\
			
			DSAGL (Ours)
			& \textcolor{red}{\textbf{0.5739 \ensuremath{\pm} 0.0087}}
			& \textcolor{red}{\textbf{0.5186 \ensuremath{\pm} 0.0102}}
			& \textcolor{red}{\textbf{0.3914 \ensuremath{\pm} 0.0118}} \\
			
			\bottomrule
			
		\end{tabular}
		
	\end{table}
	
	\section{Discussion}
	\rev{\subsection{Strengths of the Proposed Framework}}
	\rev{The proposed DSAGL framework demonstrates several strengths in the context of weakly supervised WSI classification. By introducing a dual-stream architecture, the model effectively bridges the supervision gap between bag-level and instance-level learning. The alternating training strategy facilitates bidirectional knowledge transfer between the teacher and student streams, enhancing prediction consistency and training stability. In addition, the integration of a Mamba-based encoder enables efficient modeling of long-range dependencies across image patches while maintaining linear computational complexity. Experimental results across multiple datasets confirm that DSAGL consistently outperforms state-of-the-art MIL approaches, particularly in terms of instance-level precision and boundary localization accuracy.}
	
	\rev{Moreover, the visualization analysis further validates the interpretability of DSAGL. The attention heatmaps generated by the dual-stream architecture are well aligned with diagnostic lesion regions annotated by pathologists, demonstrating that the model effectively captures clinically meaningful morphological patterns. This spatial correspondence between attention responses and pathological structures enhances model transparency and provides intuitive evidence for its potential to assist pathologists in clinical diagnosis and decision support.}
	
	\rev{\subsection{Limitations and Challenges under Weak Supervision}}
	\rev{Despite these advantages, several limitations should be acknowledged. One important challenge arises from the nature of clinical slide annotations, which are typically provided at the case level rather than at precise region-level granularity. This inevitably introduces label noise. Under such circumstances, the pseudo labels generated by the teacher stream in the early training stages may deviate from true lesion regions, potentially affecting the convergence and reliability of the student stream. Future work could incorporate noise-robust optimization, confidence-aware pseudo-label filtering, or uncertainty modeling to enhance robustness against unreliable annotations.}
	
	\rev{Beyond annotation noise, class imbalance is another inherent challenge in weakly supervised WSI classification, where positive instances are often sparse compared to a large number of normal tissue patches. In this work, we primarily address this issue through a class-weighted binary cross-entropy loss, which allows explicit rebalancing during optimization without altering the original instance distribution within each bag. We did not adopt instance-level oversampling or undersampling strategies. In MIL-based pathology settings, instance labels are latent and partially noisy, especially in early training stages when pseudo labels are still evolving. Oversampling positive instances may amplify noisy pseudo-labels and lead to overfitting, while undersampling negative instances may disrupt the contextual integrity of the bag and weaken global semantic modeling. Moreover, since pseudo labels in DSAGL are dynamically refined through alternating teacher--student optimization, static sampling strategies are less compatible with the training dynamics. Therefore, loss-level reweighting offers a more stable and flexible mechanism for handling class imbalance in our framework. Exploring adaptive sampling or curriculum-based rebalancing strategies under uncertainty-aware pseudo supervision remains an interesting direction for future work.}
	
	\rev{To further analyze the behavior and reliability of pseudo supervision under weak labels, we conduct a pseudo-label analysis based on existing experimental evidence. In settings where instance-level ground truth is available, namely CIFAR-10 and NCT-CRC, we evaluate the correlation between teacher-generated pseudo labels and ground-truth instance annotations using instance-level AUC. This metric reflects ranking-based discriminative consistency and is well suited for assessing attention-derived pseudo supervision. Although explicit confidence metrics such as entropy are not reported, the evolution of pseudo-label reliability can be indirectly inferred from the steadily increasing instance-level AUC of the teacher stream and the stable convergence behavior of the student network, which relies entirely on these pseudo labels for supervision. These observations suggest that the pseudo labels become progressively more reliable during training, rather than remaining noisy or unstable.}
	
	\rev{In our implementation, a global min--max normalization (NormProb) is adopted to map attention scores to probabilistic pseudo labels while preserving relative instance saliency. While alternative normalization strategies may influence pseudo-label distributions, a systematic sensitivity analysis of different normalization schemes is left for future work. Finally, comparisons with simpler pseudo-labeling baselines, such as random or hard pseudo labels, result in substantial performance degradation, whereas the proposed soft pseudo labels approach the upper bound achieved using ground-truth instance labels. This comparison further justifies the additional complexity of the proposed attention-guided pseudo-labeling mechanism.}
	
	\rev{\subsection{On the Use of CIFAR-10 and Evaluation Strategy}}
	\rev{From a methodological perspective, although CIFAR-10 is employed to simulate weakly supervised learning conditions, it remains a natural image dataset whose visual semantics and structural characteristics differ substantially from those of real histopathological images. Accordingly, CIFAR-10 is not treated in this work as a surrogate for pathological image analysis, but rather as a highly controllable synthetic testbed for examining optimization behavior, training stability, and pseudo-label evolution under extreme weak-supervision settings. The pathology-specific capabilities of the proposed framework—such as morphological discrimination and spatial reasoning—are instead primarily validated on the medium-resolution NCT-CRC-HE-100K dataset and the gigapixel-scale TCGA-Lung whole-slide images. Through this progressive evaluation strategy, spanning from synthetic benchmarks to real-world pathological datasets, we partially bridge the gap between controlled experimental analysis and clinical applicability. Nevertheless, further validation on additional pathology-specific weakly supervised datasets, covering a broader range of organs and disease types, remains an important direction for future work to more comprehensively assess the generalizability of the proposed approach.}
	
	\rev{\subsection{Future Directions and Clinical Deployment}}
	\rev{From a task formulation perspective, the current study mainly focuses on single-label or multi-class classification. However, in real clinical practice, a single WSI may simultaneously contain multiple co-existing pathological patterns or mixed lesion subtypes. Extending DSAGL to a multi-label or hierarchical classification framework would provide greater diagnostic flexibility and better reflect the complexity of real-world pathology. This could be achieved by introducing multi-stream output layers or sigmoid-based prediction heads to enable the simultaneous identification of multiple pathological phenotypes, thereby improving clinical interpretability and applicability.}
	
	\rev{From a deployment standpoint, although the VSSMamba encoder is lightweight in design, deploying DSAGL in clinical workflows still presents practical challenges. Processing gigapixel WSIs requires substantial computational and memory resources, which may limit real-time inference in hospital environments. Future work could explore inference acceleration, distributed patch processing, and hardware-aware model compression to improve computational efficiency and deployability. In addition, integrating the model into clinical digital pathology systems, such as PACS or LIS, while ensuring data privacy and interpretability, represents another important step toward real-world application.}
	
	\rev{Overall, DSAGL is well suited for high-resolution, structurally complex, and weakly annotated pathological images, demonstrating strong generalization across diverse data distributions. Future research will focus on further enhancing the dual-stream learning paradigm by designing more adaptive feature extractors and incorporating uncertainty-aware objectives to improve model robustness and interpretability. Moreover, extending DSAGL to multi-label recognition and achieving seamless integration with clinical workflows will pave the way toward a trustworthy and deployable AI-driven pathology analysis system.}

	\section{Conclusion}
	In this work, we propose DSAGL, a novel \rev{dual-streams} teacher-student framework for weakly supervised WSI classification. By integrating attention-guided pseudo-labeling and an alternating training strategy, DSAGL effectively bridges the gap between coarse-grained supervision and fine-grained instance-level learning. The incorporation of a Mamba-based encoder enables efficient modeling of long-range dependencies with linear complexity, allowing the model to process large-scale instance sets while preserving global context. Additionally, the FASA module enhances the model's ability to focus on sparse yet diagnostically significant regions through multi-scale attention. Experimental results on three benchmark datasets demonstrate that DSAGL consistently outperforms existing MIL-based approaches in both instance-level and bag-level classification tasks, particularly under challenging conditions such as low positive-instance ratios and heterogeneous lesion patterns. These findings highlight the effectiveness, scalability, and generalization potential of DSAGL, making it a promising solution for real-world computational pathology applications. \rev{In future work, we plan to explore more advanced teacher updating mechanisms, such as EMA-based strategies, to further enhance training stability and robustness under weak supervision.}
	
	\section*{Declaration of competing interest}
	
	The authors declare that they have no known competing financial interests or personal relationships that could have appeared to influence the work reported in this paper.
	
	\section*{Acknowledgments}
	
	\rev{This work was supported by National Natural Science Foundation of China under Grant No. 62572339 and Grant No. 61901292, The Natural Science Foundation of Shanxi Province, China under Grant No. 202303021211082, and The Graduate Scientific Research and Innovation Project of Shanxi Province under Grant No. RC2400005593.}

	\rev{We sincerely thank the reviewers and the editor for their constructive comments and suggestions.}
	
	\section*{Data availability}
	
	Data will be made available on request.
	
	%\bibliographystyle{cas-model2-names}
	\bibliographystyle{elsarticle-harv}
	\bibliography{reference}

\end{document}